\pdfoutput=1
\documentclass[11pt]{article}

\usepackage{acl}

\usepackage{times}
\usepackage{latexsym}

\usepackage[T1]{fontenc}
\usepackage[utf8]{inputenc}

\usepackage{microtype}

\usepackage{comment}
\usepackage{multirow}
\usepackage{booktabs}

\usepackage{subcaption}

\usepackage{bm}
\usepackage{bbm}
\usepackage{amsmath}
\usepackage{amssymb}
\usepackage{amsfonts}

\usepackage{mathtools}

\DeclarePairedDelimiterX{\infdivx}[2]{(}{)}{%
  #1\;\delimsize\|\;#2%
}
\newcommand{\infdiv}{\mathrm{KL}\infdivx}

\usepackage{color}
\usepackage{soul}
\usepackage{ulem}

\newcommand{\argmax}{\mathop{\mathrm{argmax}}\limits}

\providecommand{\abs}[1]{\lvert#1\rvert}

\usepackage{pifont}
\newcommand{\cmark}{\textcolor{green!80!black}{\ding{51}}}
\newcommand{\xmark}{\textcolor{red}{\ding{55}}}

\title{Making Attention Mechanisms More Robust and Interpretable \\ with Virtual Adversarial Training}

\author{
    Shunsuke Kitada \and Hitoshi Iyatomi \\
    Department of Applied Informatics, Graduate School of Science and Engineering \\ 
    Hosei University, Tokyo, Japan \\ 
    \texttt{\{shunsuke.kitada.8y@stu., iyatomi@\}hosei.ac.jp} \\
}

\begin{document}
\maketitle
\begingroup\def\thefootnote{*}\footnotetext{
Initial date submitted 19 May 2022, accepted 27 October 2022, published 28 November 2022. 
This version of the article has been accepted for publication, after peer review (when applicable) and is subject to Springer Nature’s \href{https://www.springernature.com/gp/open-research/policies/accepted-manuscript-terms}{AM terms of use}, but is not the Version of Record and does not reflect post-acceptance improvements, or any corrections. 
The Version of Record is available online at: \url{https://link.springer.com/article/10.1007/s10489-022-04301-w}
}\endgroup

\begin{abstract}
Although attention mechanisms have become fundamental components of deep learning models, they are vulnerable to perturbations, which may degrade the prediction performance and model interpretability. 
Adversarial training (AT) for attention mechanisms has successfully reduced such drawbacks by considering adversarial perturbations. 
However, this technique requires label information, and thus, its use is limited to supervised settings. 
In this study, we explore the concept of incorporating virtual AT (VAT) into the attention mechanisms, by which adversarial perturbations can be computed even from unlabeled data.
To realize this approach, we propose two general training techniques, namely VAT for attention mechanisms (Attention VAT) and ``interpretable'' VAT for attention mechanisms (Attention iVAT), which extend AT for attention mechanisms to a semi-supervised setting. 
In particular, Attention iVAT focuses on the differences in attention; thus, it can efficiently learn clearer attention and improve model interpretability, even with unlabeled data. 
Empirical experiments based on six public datasets revealed that our techniques provide better prediction performance than conventional AT-based as well as VAT-based techniques, and stronger agreement with evidence that is provided by humans in detecting important words in sentences. 
Moreover, our proposal offers these advantages without needing to add the careful selection of unlabeled data. 
That is, even if the model using our VAT-based technique is trained on unlabeled data from a source other than the target task, both the prediction performance and model interpretability can be improved.

\end{abstract}

\section{Introduction}
    \begin{figure}[t]
    \centering
    \includegraphics[width=\linewidth]{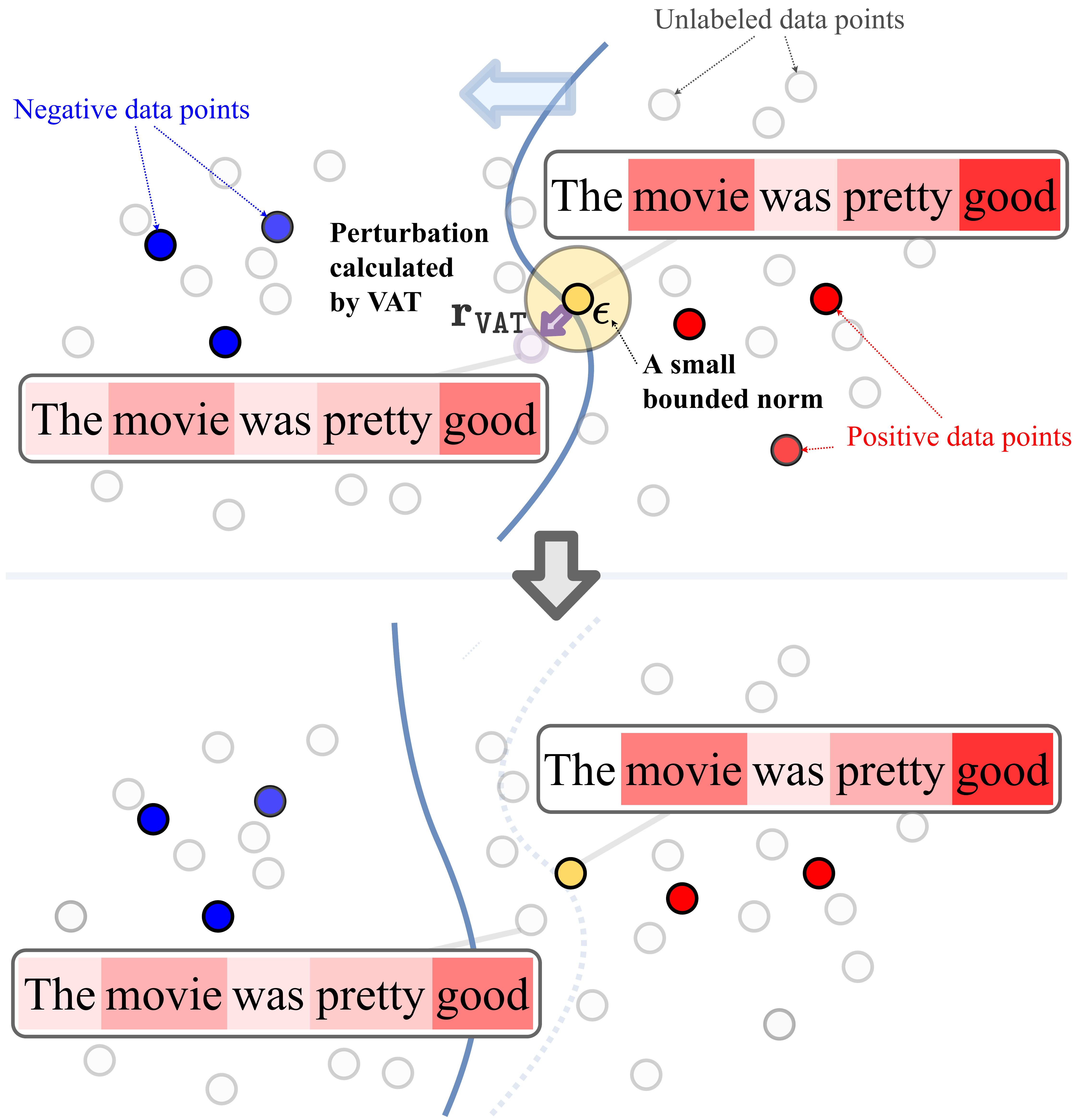}
    \caption{
        Intuitive illustration of the proposed VAT for attention mechanisms.
        Our technique can learn clearer attention by overcoming adversarial perturbations $\bm{r}_{\texttt{VAT}}$, thereby improving model interpretability
    }
    \label{fig:figure1}
\end{figure}

Despite their significant success in various computer vision (CV) and natural language processing (NLP) tasks, deep neural networks (DNNs) are usually vulnerable to tiny input perturbations, which are often referred to as adversarial examples~\cite{szegedy2013intriguing, goodfellow2014explaining, mudrakarta2018did}.
\citet{goodfellow2014explaining} proposed adversarial training (AT) for protection against malicious perturbations, which improves robustness by smoothing the discrimination boundary through adding adversarial perturbations to the input.
This technique has been reported as highly effective for solving common problems in the CV field.
In NLP, \citet{miyato2016adversarial} proposed AT for word embeddings (Word AT) to improve model robustness.
Moreover, \citet{sato2018interpretable} developed interpretable AT for word embeddings (Word iAT), which makes adversarial perturbations more interpretable by restricting the perturbation direction to existing words in the word embedding space.

At present, attention mechanisms, on which various studies have focused in recent years~\cite{bahdanau2014neural, lin2017structured, vaswani2017attention, devlin2019bert}, are among the most important components in DNN models.
However, \citet{jain2019attention} reported a critical problem; attention mechanisms are vulnerable to malicious perturbations.
To mitigate this issue, we recently proposed AT for attention mechanisms (Attention AT) and interpretable AT for attention mechanisms (Attention iAT)~\cite{kitada2021attention}.
These proposals provide model robustness to the vulnerability of attention mechanisms, as well as a more interpretable attention heatmap; that is, our techniques provide clear attention in a sentence. 
Moreover, these techniques significantly improve performance.

Although AT successfully improves model robustness, the training technique is only available in supervised settings because labeled data are needed to calculate adversarial perturbation.
However, in NLP, labeled data for training are often expensive, and it is generally not easy to apply data augmentation.
Therefore, extending these techniques to semi-supervised learning, whereby unlabeled data can also be used for training, is desirable. 
Virtual AT (VAT)~\cite{miyato2018virtual}, which is a sophisticated semi-supervised learning technique, has been proposed to extend AT to semi-supervised settings.
This technique improves  model robustness by smoothing out discrimination boundaries by calculating the adversarial perturbations, which is known as ``virtual adversarial perturbation,'' based on the current prediction using both labeled and unlabeled data.
VAT is reportedly effective in both the CV~\cite{goodfellow2014explaining, miyato2018virtual} and NLP~\cite{miyato2016adversarial, sato2018interpretable, chen2020seqvat} fields.
Semi-supervised learning methods generally use unlabeled data based on the model's output (i.e., current parameters) at the time; therefore, to a certain extent, the unlabeled data should be similar to the labeled data.

In this study, we propose two new general training techniques: \textit{VAT for attention mechanisms} (Attention VAT) and \textit{interpretable VAT for attention mechanisms} (Attention iVAT), which extend AT for attention mechanisms to a semi-supervised setting, thereby making the model even more robust and interpretable. 
Attention VAT and iVAT improve model robustness by smoothing out the discriminative boundary even for unlabeled data points by calculating the direction of the virtual adversarial perturbation.
In contrast to Attention VAT, Attention iVAT utilizes the difference in attention to each word in a sentence to construct adversarial perturbations, and can thus learn clear attention more efficiently, which improves the model interpretability.

We compared our proposed techniques with several other state-of-the-art AT-based techniques using six common NLP benchmarks. 
We evaluated the extent to which the attention weights that were obtained using our techniques agreed with the gradient-based word importance~\cite{simonyan2013deep} and rationales provided by humans~\cite{deyoung2019eraser}.
We investigated the effect on the performance of the unlabeled data being added to the training.
This empirically confirmed that our techniques could improve both the prediction performance and model interpretability, even when trained on unlabeled data from a source different from that of the target task.

The advantages of our VAT techniques for attention mechanisms are summarized as follows:
\begin{itemize}
    \item The VAT for attention mechanisms is an extension to semi-supervised learning of the effective AT for attention mechanisms, which provides improved prediction performance compared to conventional AT-based techniques and recent VAT-based techniques in semi-supervised settings.
    \item The VAT for attention mechanisms also favors model interpretability, exhibiting a stronger correlation with word importance based on gradients and higher agreement with the results of human annotations. This promising outcome has been particularly important for deep learning models in recent years.
    \item The benefit of the above extension to semi-supervised learning is almost independent of selecting the additional unlabeled data to be used. That is, a significant performance improvement can be expected even while reducing the difficulty of selecting unlabeled data. Our empirical experiments confirm that prediction performance improves by increasing the unlabeled data.
\end{itemize}

\section{Related Work}
    In this section, we describe two topics that are closely related to our VAT for attention mechanisms: (1) AT and its extension, namely VAT, and (2) attention mechanisms.

\subsection{AT and its Extension, VAT}

AT~\cite{szegedy2013intriguing, goodfellow2014explaining, wang2016theoretical, li2021adversarial} is a powerful regularization technique in the CV field primarily explored to improve model's robustness to input perturbations.
A recent proposal in the CV field, attention transfer based adversarial training (ATAT)~\cite{li2021adversarial}, considers the class activation map (CAM) as an attention heatmap and applies adversarial training to the CAM.
Furthermore, AT has been applied to various NLP tasks by extending the concept of adversarial perturbations, such as text classification~\cite{miyato2016adversarial, sato2018interpretable, yi2022fine}, part-of-speech tagging~\cite{yasunaga2018robust}, relation extraction~\cite{wu2017adversarial}, and machine reading comprehension~\cite{wang2016attention}.
The recently proposed stability fine-tuning framework (StaFF)~\cite{yi2022fine} applies AT to defend against word-level adversarial attacks.
Word AT~\cite{miyato2016adversarial} is a pioneering technique that applies AT for word embeddings in NLP.
Word iAT~\cite{sato2018interpretable} has subsequently been proposed to realize more ``interpretable'' AT for word embeddings. 
This technique restricts the perturbation direction to existing words in the word embeddings space. 
Each input with a perturbation can be interpreted as an actual sentence by considering the perturbation as a replacement for a word in the sentence. 
However, although this approach enhances the interpretability of the perturbations, it does not increase the interpretability of model prediction. 

VAT~\cite{miyato2018virtual} extends AT to semi-supervised settings,  in which \textit{virtual} adversarial perturbation, even for unlabeled data points, is defined.
This technique provides smoother discriminative boundaries and improves model robustness compared to AT, in which only supervised data are used.
VAT has achieved state-of-the-art performance in the CV field, e.g., in image classification tasks~\cite{miyato2018virtual}, as well as excellent results in the NLP fields, e.g., in text classification~\cite{miyato2016adversarial, sato2018interpretable} and sequence tagging~\cite{chen2020seqvat}.
In particular, \citet{chen2020seqvat} used an extremely large dataset, namely the One Billion Word Language Model Benchmark dataset~\cite{chelba2014one}, as an unlabeled data pool for semi-supervised learning.
Their VAT-based technique outperformed state-of-the-art sequence labeling models.
With a small number of labeled examples, \citet{an2022attention} recently proposed a VAT-based semi-supervised question generation (QG) model, named virtual stroke recognition copy network (VSAC Net), for the Chinese QG tasks.
These models that use VAT for NLP tasks are applied to input or hidden representations, which is different from our proposed concept of applying VAT to attention mechanisms.

Whereas previous AT and VAT techniques have been applied to the input space mainly for word embeddings, we proposed the Attention AT and Attention iAT techniques for the application of AT to attention mechanisms~\cite{kitada2021attention}.
These techniques outperformed Word AT and Word iAT in various tasks, such as text classification, question answering (QA), and natural language inference (NLI) tasks.
The results indicated that AT for attention mechanisms is more efficient than AT for word embeddings regarding both model accuracy and interpretability.
Our proposed Attention VAT and Attention iVAT, are semi-supervised learning extensions of Attention AT and Attention iAT, respectively.

\begin{table*}[t]
\centering
\begin{tabular}{@{}lccc@{}}
\toprule
                                         & \begin{tabular}[c]{@{}c@{}}Attention\\ vulnerability\end{tabular} & \begin{tabular}[c]{@{}c@{}}Unlabeled\\ data\end{tabular} & \begin{tabular}[c]{@{}c@{}}Model\\ interpretability\end{tabular} \\ \cmidrule(r){1-1} \cmidrule(lr){2-2} \cmidrule(lr){3-3} \cmidrule(l){4-4}
Word AT~\cite{miyato2016adversarial}     & \xmark                                                            & \xmark                                                   & \xmark                                                           \\
Word iAT~\cite{sato2018interpretable}    & \xmark                                                            & \xmark                                                   & \cmark                                                           \\ \cmidrule(r){1-1} \cmidrule(lr){2-2} \cmidrule(lr){3-3} \cmidrule(l){4-4}
Word VAT~\cite{miyato2018virtual}        & \xmark                                                            & \cmark                                                   & \xmark                                                           \\
Word iVAT~\cite{sato2018interpretable}   & \xmark                                                            & \cmark                                                   & \cmark                                                           \\ \cmidrule(r){1-1} \cmidrule(lr){2-2} \cmidrule(lr){3-3} \cmidrule(l){4-4} \morecmidrules \cmidrule(r){1-1} \cmidrule(lr){2-2} \cmidrule(lr){3-3} \cmidrule(l){4-4}
Attention AT~\cite{kitada2021attention}  & \cmark                                                            & \xmark                                                   & \xmark                                                           \\
Attention iAT~\cite{kitada2021attention} & \cmark                                                            & \xmark                                                   & \cmark                                                           \\ \cmidrule(r){1-1} \cmidrule(lr){2-2} \cmidrule(lr){3-3} \cmidrule(l){4-4} 
\textbf{Attention VAT (ours)}            & \cmark                                                            & \cmark                                                   & \xmark                                                           \\
\textbf{Attention iVAT (ours)}           & \cmark                                                            & \cmark                                                   & \cmark                                                           \\ \bottomrule
\end{tabular}%
\caption{
    Comparison of the proposed Attention VAT/iVAT with that of related studies. 
    Both of our techniques remedy the potential vulnerability to perturbations in the attention mechanisms and can handle unlabeled data. Furthermore, Attention iVAT learns clearer attention efficiently and improves the model interpretability without the selection of unlabeled data.
}
\label{tab:related_work}
\end{table*}

Table~\ref{tab:related_work} summarizes the comparison between our Attention VAT/iVAT and those in related studies. 
Both of our proposals, Attention VAT and Attention iVAT, remedy the potential vulnerability to perturbations in the attention mechanisms and can handle unlabeled data. 
Moreover, Attention iVAT efficiently learns clearer attention and improves model interpretability without the selection of unlabeled data.
VAT has often been employed for inputs and their embeddings~\cite{miyato2018virtual, sato2018interpretable}; however, to our knowledge, no application to attention mechanisms has been reported.
We have empirically demonstrated that our VAT-based techniques are effective even when they are applied to unlabeled data from sources other than the labeled data. 

\subsection{Attention Mechanisms}

Attention mechanisms for machine translation were first introduced by \citet{bahdanau2014neural}.
These mechanisms contribute to improving the prediction performance of various NLP tasks, such as sentence-level classification~\cite{lin2017structured}, sentiment analysis~\cite{wang2016attention, dai2022sentatn}, QA~\cite{he2016character}, and NLI ~\cite{parikh2016decomposable}.
The recently proposed sentence-level attention transfer network (SentATN)~\cite{dai2022sentatn} aims to solve the cross-domain sentence classification task with the advantages of the attention mechanism and sentence-level AT.

Attention weights are claimed to offer insights into the inner workings of DNNs~\cite{li2016understanding}.
However, learned attention weights are often uncorrelated with word importance, which is calculated using the gradient-based method~\cite{simonyan2013deep}, and perturbations to the attention mechanisms may interfere with the interpretation~\cite{jain2019attention, pruthi2019learning}.
Our recently proposed AT for attention mechanisms~\cite{kitada2021attention}, is a novel and practical training technique for solving these issues. 
We have demonstrated that this technique (1) improves the performance, (2) exhibits a stronger correlation with gradient-based word importance, and (3) is substantially less dependent on the scale of perturbation.

This study presents a new attempt to employ VAT in attention mechanisms.
Our proposal addresses the potential problem with attention mechanisms, namely the low correlation between the attention weight and word importance.
It is expected to increase the correlation (i.e., model interpretability) further by effectively using unlabeled data and learning to focus on the differences in the attention weights.

\section{VAT for Attention Mechanisms}
    Our proposal is an extension of training techniques that are effective by introducing adversarial perturbation to the attention mechanism to semi-supervised learning using VAT~\cite{miyato2018virtual}.
In this section, we introduce two new general training techniques for attention mechanisms based on VAT: (1) VAT for attention mechanisms (Attention VAT), and (2) interpretable VAT for attention mechanisms (Attention iVAT).
These techniques follow the calculation of the perturbation direction in the original VAT, and thus, the formulas are similar although the implications are different, as we will explain in the following sections.

Fig.~\ref{fig:figure1} provides an intuitive illustration of the proposed technique.
Our proposals provide general-purpose robust training techniques that can be applied to any model that uses attention mechanisms and can also use unlabeled data. 
We apply the proposed techniques to the common model that is described in Appendix~\ref{sec:common_model}.

Let $X_{\tilde{\bm{a}}}$ be an input sequence with an attention score $\tilde{\bm{a}}$. 
We model the conditional probability of class $\bm{y}$ as $p(\bm{y} \vert X_{\tilde{\bm{a}}}; \bm{\theta})$, where $\bm{\theta}$ represents all model parameters.
We minimize the following negative log-likelihood as a loss function for the model parameters to train the model:
\begin{equation}
    \mathcal{L}(X_{\tilde{\bm{a}}}, \bm{y}; \bm{\theta}) = - \log{p(\bm{y} \vert X_{\tilde{\bm{a}}}; \bm{\theta})}.
\end{equation}
Our proposed techniques introduce an adversarial perturbation to the attention score $\tilde{{\bm{a}}}$.

\subsection{Attention VAT}

The first proposal, Attention VAT is a natural extension of our previous Attention AT~\cite{kitada2021attention} for supporting semi-supervised learning.
Whereas AT determines the adversarial perturbation in supervised setting based on label information, VAT computes the \textit{virtual} adversarial perturbation even for unlabeled data, and is expected to provide more robust models.
The virtual adversarial perturbation on the attention mechanisms is defined as the worst-case perturbation on the mechanisms of a small, bounded norm by maximizing the Kullback–Leibler (KL) divergence between the estimated label distribution of the original examples and the distribution of the perturbation that is added to the perturbation of the original example.

Suppose that $\mathcal{D}'$ denotes a set of $N_l$ labeled and $N_u$ unlabeled data. 
We introduce $X_{\tilde{\bm{a}} + \bm{r}}$, which denotes $X_{\tilde{\bm{a}}}$ with the additional perturbation $\bm{r}$.
Attention VAT minimizes the following loss function for estimating the loss of the virtual adversarial perturbation $\bm{r}_{\tt VAT}$ on the input sequence with an attention score:
\begin{multline}
    \mathcal{L}_{\tt VAT}(X_{\tilde{\bm{a}}}, X_{\tilde{\bm{a}} + \bm{r}_{\tt VAT}}; \hat{\bm{\theta}}) \\ = \frac{1}{\abs{\mathcal{D}'}} \sum_{X \in \mathcal{D}'} \mathcal{L}_{\tt KL}(X_{\tilde{\bm{a}}}, X_{\tilde{\bm{a}} + \bm{r}_{\tt VAT}}; \hat{\bm{\theta}}),
\end{multline}
\begin{equation}
    \bm{r}_{\tt VAT} = \argmax_{\bm{r}: \|\bm{r}\|_2 \le \epsilon} \mathcal{L}_{\tt KL}(X_{\tilde{\bm{a}}}, X_{\tilde{\bm{a}} + \bm{r}}; \hat{\bm{\theta}}),\label{eq:argmax_vat}
\end{equation}
where $\epsilon$ is a hyperparameter that controls the norm of the perturbation and is determined using the validation set. $\hat{\bm{\theta}}$ represents the current model parameters.
In this case, $\mathcal{L}_{\tt KL}(X_{\tilde{\bm{a}}}, X_{\tilde{\bm{a}} + \bm{r}}; \hat{\bm{\theta}})$ is the KL divergence, which can be calculated using $\mathrm{KL}(\cdot\|\cdot)$:
\begin{multline}
    \mathcal{L}_{\tt KL}(X_{\tilde{\bm{a}}}, X_{\tilde{\bm{a}} + \bm{r}_{\tt VAT}}; \hat{\bm{\theta}}) \\ = \infdiv{p(\cdot \vert X_{\tilde{\bm{a}}}, \hat{\bm{\theta}})}{p(\cdot \vert X_{\tilde{\bm{a}} + \bm{r}_{\tt VAT}}, \hat{\bm{\theta}})}.
\end{multline}
VAT uses the current model parameters $\hat{\bm{\theta}}$ to find the worst perturbation, $\bm{r}_{\tt VAT}$, that most affects the output probability and learns, such that the prediction does not change when $\bm{r}_{\tt VAT}$ is added to each sample. 
Differently expressed, the effect is to make the model robust by smoothing the probability distribution of predictions for each data point, including the unlabeled data.
We use an approximation method instead of solving the expensive optimization problem above~\cite{miyato2016adversarial}:
\begin{equation}
    \bm{r}_{\tt VAT} = \epsilon \frac{\bm{g}}{\|\bm{g}\|_2}, \mathrm{where}~ \bm{g} = \nabla_{\tilde{\bm{a}}} \mathcal{L}_{\tt KL}(X_{\tilde{\bm{a}}}, X_{\tilde{\bm{a}} + \bm{r}}; \hat{\bm{\theta}})\label{eq:vat_derivation}.
\end{equation}
In each training step, we obtain a virtual adversarial perturbation $\bm{r}_{\tt VAT}$ for the attention score $\tilde{\bm{a}}$ of each input data based on the current model $\hat{\bm{\theta}}$:
\begin{equation}
    \tilde{\bm{a}}_{\tt vadv} = \tilde{\bm{a}} + \bm{r}_{\tt VAT}.\label{eq:alpha_plus_r_at}
\end{equation}

\subsection{Attention iVAT}\label{sec:method_attention_ivat}
The other proposal, Attention iVAT, is also an extension of our previous Attention iAT~\cite{kitada2021attention} for semi-supervised learning.
Our former Attention iAT generates adversarial perturbations by focusing on the difference in attention to each word, and thus, generates clearer differences in attention among words. 
That is, this technique significantly enhances model interpretability by providing an intuitive and clear indication of the words in a sentence that are important.

Another technique relating to our proposal, namely Word iVAT, extends interpretable AT for the word embedding of each word to semi-supervised learning. 
This technique may improve the interpretability of perturbations by restricting the direction of the added perturbation to the known direction of the word.
Although the name of this technique is similar to that of our proposal, it does not provide explicit information regarding the important words in a sentence, and thus does not provide the high interpretability of the model as the name implies.

We formulate our Attention iVAT as follows.
We define the word attentional difference vector $\bm{d}_t \in \mathbb{R}^T$, where $T$ is the number of words in the sequence, as the attentional difference between the $t$-th word $\tilde{a}_t$ and any $k$-th word $\tilde{a}_k$ in a sentence:
\begin{equation}
    \bm{d}_t = (d_{t, k})_{k=1}^{T} = (\tilde{a}_t - \tilde{a}_k)_{k=1}^{T}.
\end{equation}
Note that the difference in the attention score to itself is 0; that is, $d_{t,t} = 0$.
By normalizing the norm of the vector, we define a normalized word attentional difference vector of the attention for the $t$-th word, as follows:
\begin{equation}
    \tilde{\bm{d}}_t = \frac{\bm{d}_t}{\|\bm{d}_t\|_2}.\label{eq:normalized_attention_distance_vector}
\end{equation}
We define the perturbation $r(\bm{w}_t)$ for the attention to the $t$-th word with the trainable parameters $\bm{w}_t = (w_{t,k})_{t=1}^{T} \in \mathbb{R}^T$ as follows:
\begin{equation}
    r(\bm{w}_t) = \bm{w}^{\top}_t \tilde{\bm{d}}_t
\end{equation}
By combining $\bm{w}_t$ for all $t$ as $\bm{w}$, we can calculate the perturbation $\bm{r}(\bm{w})$ for the sentence:
\begin{equation}
    \bm{r}(\bm{w}) = (r(\bm{w}_t))_{t=1}^{T}.
\end{equation}
Subsequently, similar to $X_{\tilde{\bm{a}} + \bm{r}}$ in Eq.~(\ref{eq:argmax_vat}), we introduce $X_{\tilde{\bm{a}} + \bm{r}(\bm{w})}$ and seek the worst-case direction of the difference vectors that maximize the loss function:
\begin{equation}
    \bm{r}_{\tt iVAT} = \argmax_{\bm{r}: \|\bm{r}\|_2 \le \epsilon} \mathcal{L}_{\tt KL}(X_{\tilde{\bm{a}}}, X_{\tilde{\bm{a}} + \bm{r}(\bm{w})}; \hat{\bm{\theta}})\label{eq:argmax_ivat}.
\end{equation}
Using the same derivation method as in Eq.~(\ref{eq:vat_derivation}) to obtain the above approximation, we introduce the following equation to calculate $\bm{r}_{\tt iVAT}$ as an extension to semi-supervised learning:
\begin{equation}
    \bm{r}_{\tt iVAT} = \epsilon \frac{\bm{g}}{\|\bm{g}\|_2}, ~~\bm{g} = \nabla_{\bm{w}} \mathcal{L}_{\tt KL}(X_{\tilde{\bm{a}}}, X_{\tilde{\bm{a}} + \bm{r}(\bm{w})}, \hat{\bm{\theta}}).
\end{equation}
Similar to Eq.~(\ref{eq:alpha_plus_r_at}), we construct a perturbated example for the attention score $\tilde{\bm{a}}$:
\begin{equation}
    \tilde{\bm{a}}_{\tt ivadv} = \tilde{\bm{a}} + \bm{r}_{\tt iVAT}.
\end{equation}

Unlike Attention VAT, Attention iVAT defines the perturbation $\bm{r}_{\texttt{iVAT}}$ as the product of a trainable parameter $\bm{w}_t$ and a normalized word attentional vector $\bm{d}_t$.
In situations in which the difference in the attention among words is relatively small (i.e.,  the model does not know which words in the sentence are important), Eq.~(\ref{eq:normalized_attention_distance_vector}) enables the computation of meaningful virtual adversarial perturbations that increase the difference in the attention between words. 
Consequently, the model is expected to obtain clearer attention to overcome the perturbations.
This process can be considered as attention difference enhancement and was described in our previous paper~\cite{kitada2021attention}.

We describe the attention difference enhancement in Attention iVAT in terms of its formula.
As the norm of the difference in the attention weight (as indicated in Eq.~(\ref{eq:normalized_attention_distance_vector})) is normalized to one, virtual adversarial perturbations for attention mechanisms can clarify these differences, particularly in the case of sentences with a small difference in the attention to each word.
Even in situations in which there is little difference in the attention between each element of $\bm{d}_t = (d_{t,1}, d_{t,2}, \cdots, d_{t,T})$, the difference is emphasized by Eq.~(\ref{eq:normalized_attention_distance_vector}).
In the case of sentences in which there is originally a difference in the clear attention to each word, the regularization effect of $\hat{\bm{d}}_t$ in Eq.~(\ref{eq:normalized_attention_distance_vector}) has almost no effect because it hardly changes the ratio.
As in our previous study~\cite{kitada2021attention}, the Attention iVAT technique enhances the usefulness of VAT when it is employed in attention mechanisms by generating effective perturbations for each word.
A notable advantage of our Attention iVAT is that this attention difference enhancement can be computed without label information.

\subsection{Model Training with VAT}
We generate a virtual adversarial perturbation based on the current model $\hat{\bm{\theta}}$ at each training step.
To this end, we define the loss function for the VAT as follows:
\begin{equation}
    \tilde{\mathcal{L}} = \underbrace{\mathcal{L}(X_{\tilde{\bm{a}}}, \bm{y}; \hat{\bm{\theta}})}_{\substack{\text{Loss from} \\ \text{unmodified examples}}} + \underbrace{\lambda \mathcal{L}_{\tt VAT}(X_{\tilde{\bm{a}}}, X_{\tilde{\bm{a}}_{\tt VADV}}; \hat{\bm{\theta}})}_{\substack{\text{Loss from} \\ \text{virtual adversarial examples}}},\label{eq:overall_loss}
\end{equation}
where $\lambda$ is the coefficient that is used to control the two-loss functions.
Note here that $X_{\tilde{\bm{a}}_{\tt VADV}}$ may $X_{\tilde{\bm{a}}_{\tt vadv}}$ for Attention VAT or $X_{\tilde{\bm{a}}_{\tt ivadv}}$ for Attention iVAT.

\section{Experiments}
    In this section, we describe the comparison models, datasets, evaluation tasks, and evaluation criteria that were used in the experiments.

\subsection{Comparison Models}

We compared Attention VAT and Attention iVAT, with conventional AT- and VAT-based training techniques.
Following previous studies~\cite{jain2019attention, kitada2021attention, meister2021sparse}, we implemented the techniques using the same recurrent neural network (RNN)-based model architecture described in Appendix~\ref{sec:common_model}. 
The proposed techniques effectively extend semi-supervised learning, a training technique that achieves high performance and generalizability by adding adversarial perturbation to the attention mechanism of the model. Therefore, we conducted comparative experiments that employed the widely used RNN-based model configurations as a baseline and evaluated them by how much performance improved.
Our proposed technique is a general-purpose robust training method that can be applied to all models. 
Therefore, in this study, instead of comparing our techniques with models that aim to improve the individual performance, we used a model with a general configuration as a baseline and conducted an evaluation to compare it with other training methods that have been recently proposed.

We followed the configurations used in the literature to ensure a fair comparison of the training techniques~\cite{jain2019attention, kitada2021attention}.
Further details are provided in Appendix~\ref{sec:implementation_details}.
We compared the following models in supervised and semi-supervised settings:

\subsubsection{Supervised Models}
% In supervised settings, we compared the following seven models:
We compared the following seven models:
\begin{itemize}
    \item \textbf{Vanilla}~\cite{jain2019attention}: A model with attention mechanisms that is trained without the use of AT- or VAT-based techniques, as described in Appendix \ref{sec:common_model}.
    \item \textbf{Word AT}~\cite{miyato2016adversarial}: Word embeddings trained with AT.
    \item \textbf{Word iAT}~\cite{sato2018interpretable}: Word embeddings trained with iAT.
    \item \textbf{Attention AT}~\cite{kitada2021attention}: Attention to word embeddings trained with AT.
    \item \textbf{Attention iAT}~\cite{kitada2021attention}: Attention to word embeddings trained with iAT.
    \item \textbf{Attention VAT} (\textbf{ours}): Attention to word embeddings trained with VAT.
    \item \textbf{Attention iVAT} (\textbf{ours}): Attention to word embeddings trained with iVAT.
\end{itemize}

\subsubsection{Semi-supervised Models}
% In semi-supervised settings, we compared the following four models:
We compared the following four models:
\begin{itemize}
    \item \textbf{Word VAT}~\cite{miyato2018virtual}: Word embeddings trained with VAT.
    \item \textbf{Word iVAT}~\cite{sato2018interpretable}: Word embeddings trained with iVAT.
    \item \textbf{Attention VAT} (\textbf{ours}): Attention to  word embeddings trained with VAT.
    \item \textbf{Attention iVAT} (\textbf{ours}): Attention to word embeddings trained with iVAT. 
\end{itemize}

\subsection{Datasets and Tasks}\label{sec:datasets_and_tasks}

Table~\ref{tab:dataset} presents the statistics for the labeled datasets, which include datasets for single-sequence tasks (e.g., text classification) and pair-sequence tasks (e.g., QA and NLI).
We divided the dataset into training, validation, and test sets as done in \cite{kitada2021attention}.
Moreover, we preprocessed these datasets according to \cite{jain2019attention, kitada2021attention}.
Refer to Appendix~\ref{sec:appendix_tasks_and_dataset} for further details on the dataset and preprocessing.

\subsubsection{Labeled Dataset} 

Following~\cite{jain2019attention, kitada2021attention}, we evaluated the techniques using open datasets for three single-sequence and three pair-sequence tasks.

\paragraph{Labeled Dataset for Single-Sequence Tasks} We used the following four datasets: Standard Sentiment Treebank (\textbf{SST})~\cite{socher2013recursive}, \textbf{IMDB}, a large movie reviews corpus~\cite{maas2011learning}, and the \textbf{AGNews} corpus~\cite{zhang2015character}.
The model was trained using the entire training set.

\paragraph{Labeled Dataset for Pair-Sequence Tasks}\label{sec:labeled_dataset_pair_sequence}

We used the following three datasets: the \textbf{CNN} news article corpus~\cite{hermann2015teaching}, The Stanford Natural Language Inference (\textbf{SNLI})~\cite{bowman2015large}, and the Multi-Genre NLI (\textbf{MultiNLI})~\cite{williams2017broad}.
The model was trained using half of the randomly sampled training data.
The other half was used as the unlabeled data, as described in Section~\ref{sec:unlabeled_dataset_pair_sequence}.

\subsubsection{Unlabeled Dataset} 
We describe the unlabeled data for the single-sequence and pair-sequence tasks.

\paragraph{Unlabeled Dataset for Single-Sequence Tasks}\label{sec:unlabeled_dataset_single_sequence}

We used the One Billion Word Language Model Benchmark dataset~\cite{chelba2014one} as an unlabeled data pool for the semi-supervised learning step.
This benchmark is used extensively to evaluate language modeling techniques and was recently employed by \citet{chen2020seqvat} for the model based on VAT.
Thus, we considered that the dataset would also be useful in our VAT-based settings.

\paragraph{Unlabeled Dataset for Pair-Sequence Tasks}\label{sec:unlabeled_dataset_pair_sequence}

Owing to the nature of the pair-sequence task, it is inherently difficult to use external sentences as unlabeled data, as in the case of single-sequence tasks.
For training, we, therefore, decided to use half of the randomly sampled training data as labeled data and the other half as unlabeled data. 
The evaluation was performed using the unlabeled data.
We adopted this evaluation method as it was used in the original VAT paper~\cite{miyato2016adversarial} and is a viable option.

\begin{table*}[t]
\centering
\resizebox{\textwidth}{!}{%
\begin{tabular}{@{}lllrrrrrrr@{}}
\toprule
\multicolumn{1}{c}{\multirow{2}{*}{Task}} & \multicolumn{1}{c}{\multirow{2}{*}{Dataset}} & \multicolumn{1}{c}{\multirow{2}{*}{}} & \multicolumn{1}{c}{\multirow{2}{*}{\# classes}} & \multicolumn{2}{c}{\# train}                                                      & \multicolumn{1}{c}{\multirow{2}{*}{\# valid}} & \multicolumn{1}{c}{\multirow{2}{*}{\# test}} & \multicolumn{1}{c}{\multirow{2}{*}{\# vocab}} & \multicolumn{1}{c}{\multirow{2}{*}{Avg. \# words}} \\
\multicolumn{1}{c}{}                      & \multicolumn{1}{c}{}                         & \multicolumn{1}{c}{}                  & \multicolumn{1}{c}{}                          & \multicolumn{1}{c}{\# labeled ($N_l$)} & \multicolumn{1}{c}{\# unlabeled ($N_u$)} & \multicolumn{1}{c}{}                         & \multicolumn{1}{c}{}                         & \multicolumn{1}{c}{}                          & \multicolumn{1}{c}{}                               \\ \cmidrule(r){1-1} \cmidrule(lr){2-3} \cmidrule(lr){4-4} \cmidrule(lr){5-5} \cmidrule(lr){6-6} \cmidrule(lr){7-7} \cmidrule(lr){8-8} \cmidrule(lr){9-9} \cmidrule(l){10-10}
Single sequence task                      & SST                                          & \cite{socher2013recursive}            & 2                                             & 6,920                                  & -                                        & 872                                          & 1,821                                        & 13,723                                        & 18                                                 \\
                                          & IMDB                                         & \cite{maas2011learning}               & 2                                             & 17.186                                 & -                                        & 4,294                                        & 4,353                                        & 12,485                                        & 171                                                \\
                                          & AGNews                                       & \cite{zhang2015character}             & 2                                             & 51,000                                 & -                                        & 9,000                                        & 3,800                                        & 13,713                                        & 35                                                 \\ \cmidrule(r){1-1} \cmidrule(lr){2-3} \cmidrule(lr){4-4} \cmidrule(lr){5-5} \cmidrule(lr){6-6} \cmidrule(lr){7-7} \cmidrule(lr){8-8} \cmidrule(lr){9-9} \cmidrule(l){10-10}
Pair sequence task                        & CNN news                                         & \cite{hermann2015teaching}              & 584                                           & 190,149                                 & 190,149                                  & 3,924                                        & 3,198                                        & 70,192                                        & 773                                                \\
                                          & SNLI                                         & \cite{bowman2015large}                & 3                                             & 274,683                                & 274,684                                  & 9,842                                        & 9,824                                        & 20,979                                        & 22                                                 \\
                                          & MultiNLI                                    & \cite{williams2017broad}              & 3                                             & 157,080                                & 157,081                                  & 78,541                                       & 19,647                                       & 53,112                                        & 34                                                 \\ \bottomrule
\end{tabular}
}
\caption{Dataset statistics. 
The unlabeled data were randomly sampled from the One Billion Word Language Model Benchmark~\cite{chelba2014one} for the single-sequence task and randomly divided into halves from the training set for the pair-sequence task.
}
\label{tab:dataset}
\end{table*}

\subsection{Evaluation Criteria}

We conducted four evaluations of model performance and interpretability when the proposed techniques were applied: (1) prediction performance following previous studies, (2) correlations based on word importance, (3) agreement with human annotations, and (4) the effects of the amount and selection of unlabeled data.

\subsubsection{Prediction Performance} 

We used the F1 score, accuracy, and micro-F1 score for the single- and pair-sequence tasks, as in \cite{jain2019attention, kitada2021attention}, to compare the performance of the model when the proposed and conventional VAT-based techniques were applied.
The models were evaluated in supervised and semi-supervised settings.

\subsubsection{Correlation with Word Importance} 

We compared how well the attention weights obtained using our VAT-based technique agreed with the importance of the words that were calculated using gradients~\cite{simonyan2013deep}. 
We used Pearson's correlations between the attention weights and the word importance, as in the work of \citet{mohankumar2020towards} and our previous research~\cite{kitada2021attention}, to evaluate the agreement. 
Refer to Appendix~\ref{sec:appendix_correlation_attention_gradient} for details on the evaluation criteria.

\subsubsection{Agreement with Human Annotation} 

We determined how strongly the attention weights that were obtained using our technique agreed with human annotations.
\citet{deyoung2019eraser} recently released a unified benchmark dataset including movie reviews, QA, and NLI, which featured human rationales for the decisions.
They included both instance-level labels and the corresponding supporting snippets (i.e., rationales) from human annotators.
We used this annotated dataset as an additional evaluation criterion for the movie review task in the IMDB dataset.

In accordance with \citet{deyoung2019eraser}, we used metrics that are appropriate for models that perform hard (discrete) rationale selection and soft (continuous) rationale selection.
We used the intersection over union (IOU) and token-level F1 score as evaluation criteria for the hard rationale selection, and the area under the precision-recall curve (AUPRC), average precision (AP), and the area under the receiver operating characteristic curve (ROC-AUC) as evaluation criteria for the soft rationale selection, as in~\citet{deyoung2019eraser}.

\subsubsection{Effects of Amount and Selection on Unlabeled Data}\label{sec:sampling_strategy}

To understand further the effects of unlabeled data in semi-supervised learning, we analyzed the impact of the amount of unlabeled data on the model's performance.
We specifically focused on VAT-based techniques, such as Word VAT, Word iVAT, and our Attention VAT/iVAT.

Moreover, we investigated the effect of the selection strategy for the unlabeled data that were used in the semi-supervised learning in each task.
We compared two strategies: the addition of unlabeled data without considering any of the contents, and the addition of data that are semantically close to the labeled data.
Specifically, we applied random selection to the former and approximate nearest neighbor (ANN)-based selection to the latter. 

\paragraph{Random Selection}
The random selection strategy randomly samples data from an unlabeled data pool.
Unlabeled data can be prepared in large quantities compared to labeled data.
Considering the relatively small size of the labeled dataset, the number of unlabeled datasets was adjusted accordingly for each task.

\paragraph{Approximate Nearest Neighbor (ANN) Selection}
The ANN selection strategy samples data from an unlabeled data pool in which the representations are close to the sentence representation that are contained in the labeled data.
In this study, we applied ANN selection based on the following sentence representation:
\begin{itemize}
    \item \textit{Bag of words (BoW)-based representation}: The BoW representation was computed for each sentence in the pool.
    \item \textit{Average word embedding-based representation}: The average of the word embeddings for each sentence in the pool was computed as the representation.
\end{itemize}

We selected the unlabeled data near $N_{\tt ANN}$ (as the hyperparameter) for these representations using the ANN strategy and used these data for semi-supervised learning.
We used Annoy\footnote{spotify/annoy: \url{https://github.com/spotify/annoy}} to search for neighbors and set $N_{\tt ANN}$ to 10 as the neighborhood in the experiments.
We designed the experiment for the ANN selection strategy to select the same amount of data as the random selection strategy.

\section{Results}
    Tables~\ref{tab:result_at_and_vat} and \ref{tab:result_iat_and_ivat} summarize the results of the prediction performance, and the correlations between the learned attention weights and word importance that was estimated using the gradient-based method, respectively.
In both tables, (a) presents the results of the AT- and VAT-based techniques in which adversarial perturbations were added to the word or attention, whereas (b) displays the results of the iAT- and iVAT-based techniques in which improved interpretability was considered.

\subsection{Prediction Performance}

\begin{table*}[t]
    \begin{minipage}{\linewidth}
        \centering
        \subcaption{AT- and VAT-based techniques}
\label{tab:result_at_and_vat_perf}
\resizebox{\textwidth}{!}{%
\begin{tabular}{@{}lcrrrrrr@{}}
\toprule
\multicolumn{1}{c}{}                    &                                                           & \multicolumn{3}{c}{Single-sequence task}                                        & \multicolumn{3}{c}{Pair-sequence task}                                                   \\ \cmidrule(r){3-5} \cmidrule(l){6-8}
\multicolumn{1}{c}{Model}               &                                                           & \multicolumn{3}{c}{F1 [\%]}                                                     & \multicolumn{1}{c}{Acc. [\%]} & \multicolumn{2}{c}{Mic. F1 [\%]}                         \\ \cmidrule(l){3-5} \cmidrule(lr){6-6} \cmidrule(l){7-8}
\multicolumn{1}{c}{}                    & \begin{tabular}[c]{@{}c@{}}Unlabeled \\ data\end{tabular} & \multicolumn{1}{c}{SST\footnotemark[1]} & \multicolumn{1}{c}{IMDB\footnotemark[2]} & \multicolumn{1}{c}{AgNews\footnotemark[3]} & \multicolumn{1}{c}{CNN\footnotemark[4]}       & \multicolumn{1}{c}{SNLI\footnotemark[5]} & \multicolumn{1}{c}{MultiNLI\footnotemark[6]} \\ \cmidrule(r){1-1} \cmidrule(lr){2-2} \cmidrule(lr){3-3} \cmidrule(lr){4-4} \cmidrule(lr){5-5} \cmidrule(lr){6-6} \cmidrule(lr){7-7} \cmidrule(l){8-8}
Vanilla~\cite{jain2019attention}        & \xmark                                                    & 79.27                   & 88.77                    & 95.27                      & 56.25                         & 73.52                    & 56.59                         \\ \cmidrule(r){1-1} \cmidrule(lr){2-2} \cmidrule(lr){3-3} \cmidrule(lr){4-4} \cmidrule(lr){5-5} \cmidrule(lr){6-6} \cmidrule(lr){7-7} \cmidrule(l){8-8}
Word AT~\cite{miyato2016adversarial}    & \xmark                                                    & 79.61                   & 89.65                    & 95.59                      & 60.87                         & 75.85                    & 57.82                         \\
Word VAT~\cite{miyato2018virtual}       & \cmark                                                    & 83.04                   & 92.21                    & 96.23                      & 63.74                         & 77.38                    & 60.43                         \\ \cmidrule(r){1-1} \cmidrule(lr){2-2} \cmidrule(lr){3-3} \cmidrule(lr){4-4} \cmidrule(lr){5-5} \cmidrule(lr){6-6} \cmidrule(lr){7-7} \cmidrule(l){8-8}
Attention AT~\cite{kitada2021attention} & \xmark                                                    & 81.72                   & 90.00                    & 96.12                      & 62.08                         & 77.21                    & 59.28                         \\
Attention VAT (\textbf{ours})           & \cmark                                                    & \textbf{83.18}          & \textbf{92.48}           & \textbf{96.35}             & \textbf{64.93}                & \textbf{77.87}           & \textbf{61.07}                \\ \bottomrule
\end{tabular}%
}
    \end{minipage}
    
    \begin{minipage}{\linewidth}
        \vspace{2mm}
    \end{minipage}

    \begin{minipage}{\linewidth}
        \centering
        \subcaption{iAT- and iVAT-based techniques}
\label{tab:result_iat_and_ivat_perf}
\resizebox{\textwidth}{!}{%
\begin{tabular}{@{}lcrrrrrr@{}}
\toprule
                                         &                                                          & \multicolumn{3}{c}{Single-sequence task}                                        & \multicolumn{3}{c}{Pair-sequence task}                                                   \\ \cmidrule(r){3-5} \cmidrule(l){6-8} 
\multicolumn{1}{c}{Model}                &                                                          & \multicolumn{3}{c}{F1 [\%]}                                                     & \multicolumn{1}{c}{Acc. [\%]} & \multicolumn{2}{c}{Mic. F1 [\%]}                         \\ \cmidrule(l){3-5} \cmidrule(lr){6-6} \cmidrule(l){7-8} 
                                         & \begin{tabular}[c]{@{}c@{}}Unlabeled\\ data\end{tabular} & \multicolumn{1}{c}{SST\footnotemark[1]} & \multicolumn{1}{c}{IMDB\footnotemark[2]} & \multicolumn{1}{c}{AgNews\footnotemark[3]} & \multicolumn{1}{c}{CNN\footnotemark[4]}       & \multicolumn{1}{c}{SNLI\footnotemark[5]} & \multicolumn{1}{c}{MultiNLI\footnotemark[6]} \\ \cmidrule(r){1-1} \cmidrule(lr){2-2} \cmidrule(lr){3-3} \cmidrule(lr){4-4} \cmidrule(lr){5-5} \cmidrule(lr){6-6} \cmidrule(lr){7-7} \cmidrule(l){8-8}
Vanilla~\cite{jain2019attention}         & \xmark                                                   & 79.27                   & 88.77                    & 95.27                      & 56.25                         & 73.52                    & 56.59                         \\ \cmidrule(r){1-1} \cmidrule(lr){2-2} \cmidrule(lr){3-3} \cmidrule(lr){4-4} \cmidrule(lr){5-5} \cmidrule(lr){6-6} \cmidrule(lr){7-7} \cmidrule(l){8-8}
Word iAT~\cite{sato2018interpretable}    & \xmark                                                   & 79.57                   & 89.64                    & 95.62                      & 61.21                         & 75.91                    & 57.91                         \\
Word iVAT~\cite{sato2018interpretable}   & \cmark                                                   & 83.07                   & 92.30                    & 96.17                      & 63.76                         & 77.42                    & 60.47                         \\ \cmidrule(r){1-1} \cmidrule(lr){2-2} \cmidrule(lr){3-3} \cmidrule(lr){4-4} \cmidrule(lr){5-5} \cmidrule(lr){6-6} \cmidrule(lr){7-7} \cmidrule(l){8-8}
Attention iAT~\cite{kitada2021attention} & \xmark                                                   & 82.20                   & 90.21                    & 96.19                      & 62.15                         & 77.23                    & 59.42                         \\
Attention iVAT (\textbf{ours})           & \cmark                                                   & \textbf{83.22}          & \textbf{92.56}           & \textbf{96.48}             & \textbf{65.08}                & \textbf{78.13}           & \textbf{61.18}                \\ \bottomrule
\end{tabular}%
}
    \end{minipage}
    
    \begin{minipage}{\linewidth}
        \vspace{2mm}
    \end{minipage}
    
    \resizebox{\textwidth}{!}{%
\begin{tabular}{@{}lll@{}}
$^1N_u = $ 50,000; $N_u/N_l = $ 7.22.  & $^2N_u = $ 150,000; $N_u/N_l = $ 8.23. & $^3N_u = $ 250,000; $N_u/N_l = $ 4.90. \\
$^4N_u = $ 190,149; $N_u/N_l = $ 1.00. & $^5N_u = $ 270,683; $N_u/N_l = $ 1.00. & $^6N_u = $ 157,081; $N_u/N_l = $ 1.00.
\end{tabular}%
}
    
    \caption{Comparison of prediction performance. Unlabeled data were also used for the results of the VAT-based method, and the amounts of such data are listed below the table.}
    \label{tab:result_at_and_vat}

\end{table*}

Table~\ref{tab:result_at_and_vat_perf} indicates that our Attention VAT outperformed adversarial training for word embedding (Word AT), its semi-supervised extension (Word VAT), and adversarial training for attention mechanisms (Attention AT) in all performance measures.
Moreover, Table~\ref{tab:result_iat_and_ivat_perf} shows that our Attention iVAT exhibited the best prediction performance for all benchmarks.
Thus, it can be confirmed that our Attention VAT/iVAT achieves better prediction performance than recent AT- and VAT-based techniques for the input space.

\subsection{Correlation with Word Importance}\label{sec:result_correlation_attention_gradient}

\begin{table*}[t]
    \begin{minipage}{\linewidth}
        \centering
        \subcaption{AT- and VAT-based techniques}
\label{tab:result_at_and_vat_corr}
\resizebox{\textwidth}{!}{%
\begin{tabular}{@{}lcrrrrrr@{}}
\toprule
                                        &                                                          & \multicolumn{3}{c}{Single-sequence task}                                        & \multicolumn{3}{c}{Pair-sequence task}                                             \\ \cmidrule(r){3-5} \cmidrule(l){6-8}
\multicolumn{1}{c}{Model}               &                                                          & \multicolumn{6}{c}{Pearson's correlation}                                                                                                                              \\ \cmidrule(l){3-8} 
                                        & \begin{tabular}[c]{@{}c@{}}Unlabeled\\ data\end{tabular} & \multicolumn{1}{c}{SST\footnotemark[1]} & \multicolumn{1}{c}{IMDB\footnotemark[2]} & \multicolumn{1}{c}{AgNews\footnotemark[3]} & \multicolumn{1}{c}{CNN\footnotemark[4]} & \multicolumn{1}{c}{SNLI\footnotemark[5]} & \multicolumn{1}{c}{MultiNLI\footnotemark[5]} \\  \cmidrule(r){1-1} \cmidrule(lr){2-2} \cmidrule(lr){3-3} \cmidrule(lr){4-4} \cmidrule(lr){5-5} \cmidrule(lr){6-6} \cmidrule(lr){7-7} \cmidrule(l){8-8}
Vanilla~\cite{jain2019attention}        & \xmark                                                   & 0.852                   & 0.788                    & 0.822                      & 0.638                   & 0.741                    & 0.517                         \\  \cmidrule(r){1-1} \cmidrule(lr){2-2} \cmidrule(lr){3-3} \cmidrule(lr){4-4} \cmidrule(lr){5-5} \cmidrule(lr){6-6} \cmidrule(lr){7-7} \cmidrule(l){8-8}
Word AT~\cite{miyato2016adversarial}    & \xmark                                                   & 0.647                   & 0.838                    & 0.813                      & 0.691                   & 0.757                    & 0.574                         \\
Word VAT~\cite{miyato2018virtual}       & \cmark                                                   & 0.779                   & 0.853                    & 0.831                      & 0.749                   & 0.758                    & 0.629                         \\  \cmidrule(r){1-1} \cmidrule(lr){2-2} \cmidrule(lr){3-3} \cmidrule(lr){4-4} \cmidrule(lr){5-5} \cmidrule(lr){6-6} \cmidrule(lr){7-7} \cmidrule(l){8-8}
Attention AT~\cite{kitada2021attention} & \xmark                                                   & 0.852                   & 0.819                    & 0.835                      & 0.742                   & 0.769                    & 0.593                         \\
Attention VAT (\textbf{ours})           & \cmark                                                   & \textbf{0.898}          & \textbf{0.879}           & \textbf{0.903}             & \textbf{0.766}          & \textbf{0.784}           & \textbf{0.638}                \\ \bottomrule
\end{tabular}%
}
    \end{minipage}
    
    \begin{minipage}{\linewidth}
        \vspace{2mm}
    \end{minipage}

    \begin{minipage}{\linewidth}
        \centering
        \subcaption{iAT- and iVAT-based techniques}
\label{tab:tab:result_iat_and_ivat_corr}
\resizebox{\textwidth}{!}{%
\begin{tabular}{@{}lcrrrrrr@{}}
\toprule
                                         &                                                             & \multicolumn{3}{c}{Single-sequence task}                                        & \multicolumn{3}{c}{Pair-sequence task}                                             \\ \cmidrule(r){3-5} \cmidrule(l){6-8}
\multicolumn{1}{c}{Model}                &                                                             & \multicolumn{6}{c}{Pearson's correlation}                                                                                                                              \\ \cmidrule(l){3-8} 
                                         & \begin{tabular}[c]{@{}c@{}}Unlabeled\\ dataset\end{tabular} & \multicolumn{1}{c}{SST\footnotemark[1]} & \multicolumn{1}{c}{IMDB\footnotemark[2]} & \multicolumn{1}{c}{AgNews\footnotemark[3]} & \multicolumn{1}{c}{CNN\footnotemark[4]} & \multicolumn{1}{c}{SNLI\footnotemark[5]} & \multicolumn{1}{c}{MultiNLI\footnotemark[6]} \\  \cmidrule(r){1-1} \cmidrule(lr){2-2} \cmidrule(lr){3-3} \cmidrule(lr){4-4} \cmidrule(lr){5-5} \cmidrule(lr){6-6} \cmidrule(lr){7-7} \cmidrule(l){8-8}
Vanilla~\cite{jain2019attention}         & \xmark                                                      & 0.852                   & 0.788                    & 0.822                      & 0.638                   & 0.741                    & 0.517                         \\ \cmidrule(r){1-1} \cmidrule(lr){2-2} \cmidrule(lr){3-3} \cmidrule(lr){4-4} \cmidrule(lr){5-5} \cmidrule(lr){6-6} \cmidrule(lr){7-7} \cmidrule(l){8-8}
Word iAT~\cite{sato2018interpretable}    & \xmark                                                      & 0.643                   & 0.839                    & 0.809                      & 0.702                   & 0.758                    & 0.581                         \\
Word iVAT~\cite{sato2018interpretable}   & \cmark                                                      & 0.781                   & 0.859                    & 0.893                      & 0.751                   & 0.761                    & 0.631                         \\ \cmidrule(r){1-1} \cmidrule(lr){2-2} \cmidrule(lr){3-3} \cmidrule(lr){4-4} \cmidrule(lr){5-5} \cmidrule(lr){6-6} \cmidrule(lr){7-7} \cmidrule(l){8-8}
Attention iAT~\cite{kitada2021attention} & \xmark                                                      & 0.876                   & 0.861                    & 0.903                      & 0.753                   & 0.772                    & 0.622                         \\
Attention iVAT (\textbf{ours})           & \cmark                                                      & \textbf{0.901}          & \textbf{0.883}           & \textbf{0.907}                      & \textbf{0.773}          & \textbf{0.808}           & \textbf{0.647}                \\ \bottomrule
\end{tabular}%
}
    \end{minipage}
    
    \begin{minipage}{\linewidth}
        \vspace{2mm}
    \end{minipage}
    
    \resizebox{\textwidth}{!}{%
\begin{tabular}{@{}lll@{}}
$^1N_u = $ 50,000; $N_u/N_l = $ 7.22.  & $^2N_u = $ 150,000; $N_u/N_l = $ 8.23. & $^3N_u = $ 250,000; $N_u/N_l = $ 4.90. \\
$^4N_u = $ 190,149; $N_u/N_l = $ 1.00. & $^5N_u = $ 270,683; $N_u/N_l = $ 1.00. & $^6N_u = $ 157,081; $N_u/N_l = $ 1.00.
\end{tabular}%
}
    
    \caption{Comparison of Pearson's correlation between attention to words and word importance estimated by gradients. The results are provided for the iVAT-based technique using unlabeled data.}
    \label{tab:result_iat_and_ivat}
\end{table*}

According to Table~\ref{tab:result_at_and_vat_corr}, the proposed Attention VAT yielded the strongest correlation between attention to words and word importance as determined by the gradients.
In particular, VAT, which enables the use of unlabeled data, significantly improved the performance.
As indicated in Table~\ref{tab:tab:result_iat_and_ivat_corr}, the proposed Attention iVAT achieved the highest correlation value in all datasets.
Although our previous Attention iAT exhibited a significant increase in performance, Attention iVAT further increased the correlation by a substantial margin.

\subsection{Agreement with Human Annotation}\label{sec:result_hard_rationale_soft_rationale}

\begin{table*}[t]

    \centering
    \begin{minipage}{\linewidth}
        \centering
\subcaption{
    Hard rationale selection
}
\begin{tabular}{@{}lrr@{}}
\toprule
\multicolumn{1}{c}{Model}                     & \multicolumn{1}{c}{IOU F1} & \multicolumn{1}{c}{Token F1} \\ \cmidrule(r){1-1} \cmidrule(lr){2-2} \cmidrule(l){3-3}
Vanilla~\cite{jain2019attention}                   & 0.011                      & 0.097                        \\ \cmidrule(r){1-1} \cmidrule(lr){2-2} \cmidrule(l){3-3}
Word VAT~\cite{miyato2018virtual}                  & 0.018                      & 0.100                        \\
Word iVAT~\cite{sato2018interpretable}                 & 0.019                      & 0.102                        \\ \cmidrule(r){1-1} \cmidrule(lr){2-2} \cmidrule(l){3-3}
Attention VAT (\textbf{ours})            & 0.030                      & 0.126                        \\
Attention iVAT (\textbf{ours})            & \textbf{0.033}             & \textbf{0.128}               \\ \cmidrule(r){1-1} \cmidrule(lr){2-2} \cmidrule(l){3-3} \morecmidrules \cmidrule(r){1-1} \cmidrule(lr){2-2} \cmidrule(l){3-3}
BERT-to-BERT~\cite{deyoung2019eraser} & 0.075                      & 0.145                        \\ \bottomrule
\end{tabular}%
\label{tab:hard_rationale}
    \end{minipage}

    \begin{minipage}{\linewidth}
        \centering
\subcaption{Soft rationale selection}
\begin{tabular}{@{}lrrr@{}}
\toprule
\multicolumn{1}{c}{Model}                     & \multicolumn{1}{c}{AUPRC} & \multicolumn{1}{c}{AP} & \multicolumn{1}{l}{ROC-AUC} \\ \cmidrule(r){1-1} \cmidrule(lr){2-2} \cmidrule(lr){3-3} \cmidrule(l){4-4}
Vanilla~\cite{jain2019attention}                   & 0.326                     & 0.395                  & 0.563                       \\ \cmidrule(r){1-1} \cmidrule(lr){2-2} \cmidrule(lr){3-3} \cmidrule(l){4-4}
Word VAT~\cite{miyato2018virtual}                  & 0.349                     & 0.413                  & 0.581                       \\
Word iVAT~\cite{sato2018interpretable}                 & 0.350                     & 0.414                  & 0.582                       \\ \cmidrule(r){1-1} \cmidrule(lr){2-2} \cmidrule(lr){3-3} \cmidrule(l){4-4}
Attention VAT (\textbf{Proposed})             & 0.403                     & 0.477                  & 0.646                       \\
Attention iVAT (\textbf{Proposed})            & \textbf{0.417}            & \textbf{0.489}         & \textbf{0.651}                       \\ \cmidrule(r){1-1} \cmidrule(lr){2-2} \cmidrule(lr){3-3} \cmidrule(l){4-4} \morecmidrules \cmidrule(r){1-1} \cmidrule(lr){2-2} \cmidrule(lr){3-3} \cmidrule(l){4-4}
BERT-to-BERT~\cite{deyoung2019eraser} & 0.502                     & -                      & -                           \\ \bottomrule
\end{tabular}%
\label{tab:soft_rationale}
    \end{minipage}
    
    \caption{
        Performance of vanilla and semi-supervised models for hard and soft rationale selection.
    }
    \label{tab:rationale}
    
\end{table*}

Table~\ref{tab:rationale} compares the estimated important words (i.e., the attention) in the sentences with human annotations.
The proposed techniques, especially our Attention iVAT, were more consistent with the human-provided rationales than existing AT techniques for the word as well as the baseline in both the hard and soft rationale selections.
The BERT-to-BERT pipeline model~\cite{deyoung2019eraser}\footnote{They constructed a simple model in which they first trained the encoder to extract rationales, and then trained the decoder to perform prediction using only rationales based on the pipeline model~\cite{lehman2019inferring}. The pipeline model adopts BERT for both the encoder and the decoder.} exhibited the strongest agreement with the human annotation. 
However, for rationale selection, it underwent supervised training with human annotations in a supervised manner for rationale selection, whereas the models that were applied in our technique did not use annotations, which means that the selection could be considered as unsupervised.

\subsection{Amount and Selection of Unlabeled Data}\label{sec:result_unlabemed_data}

\begin{figure*}
    \begin{minipage}{0.24\linewidth}
        \centering
        \includegraphics[width=\linewidth]{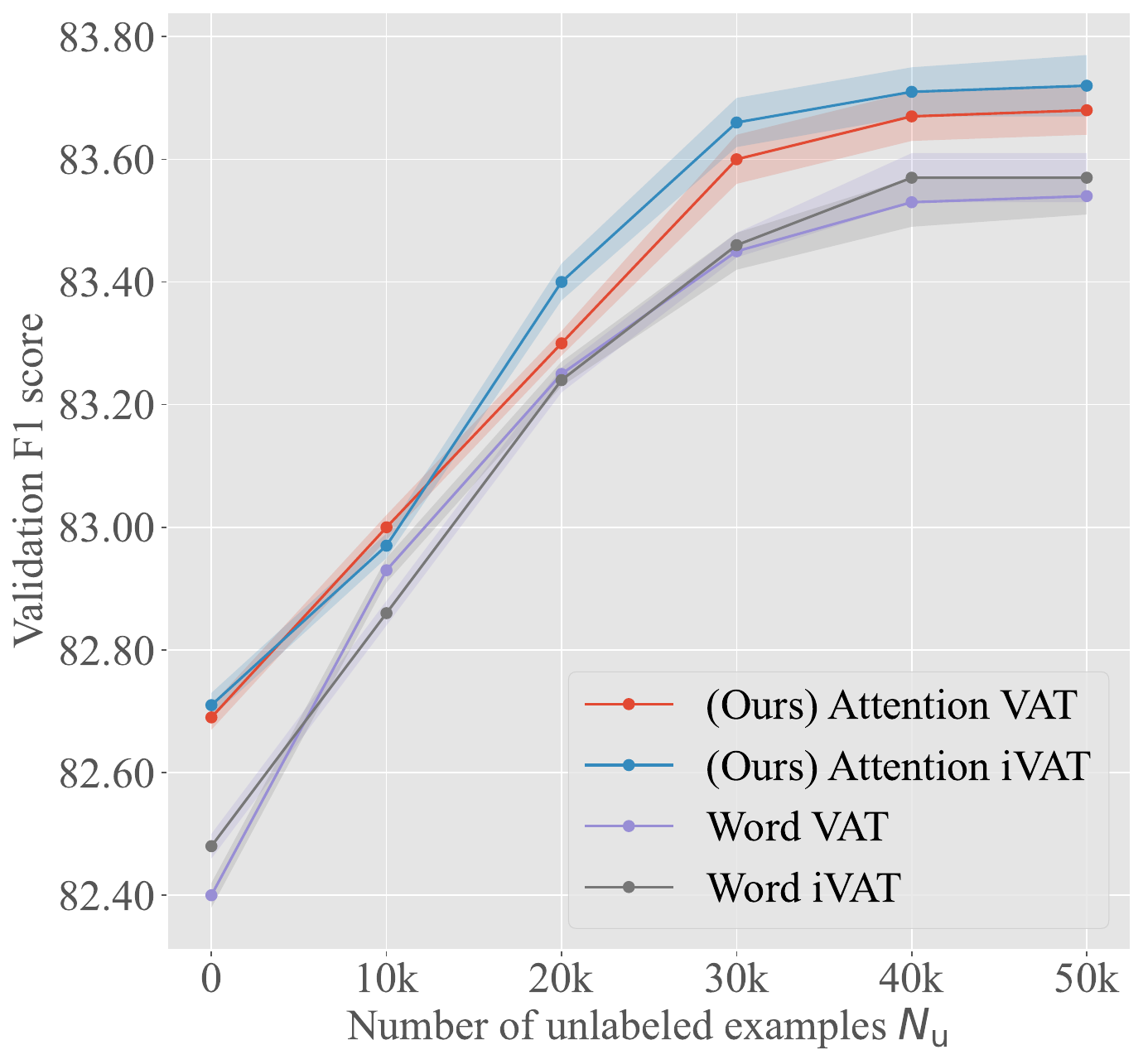}
        \includegraphics[width=\linewidth]{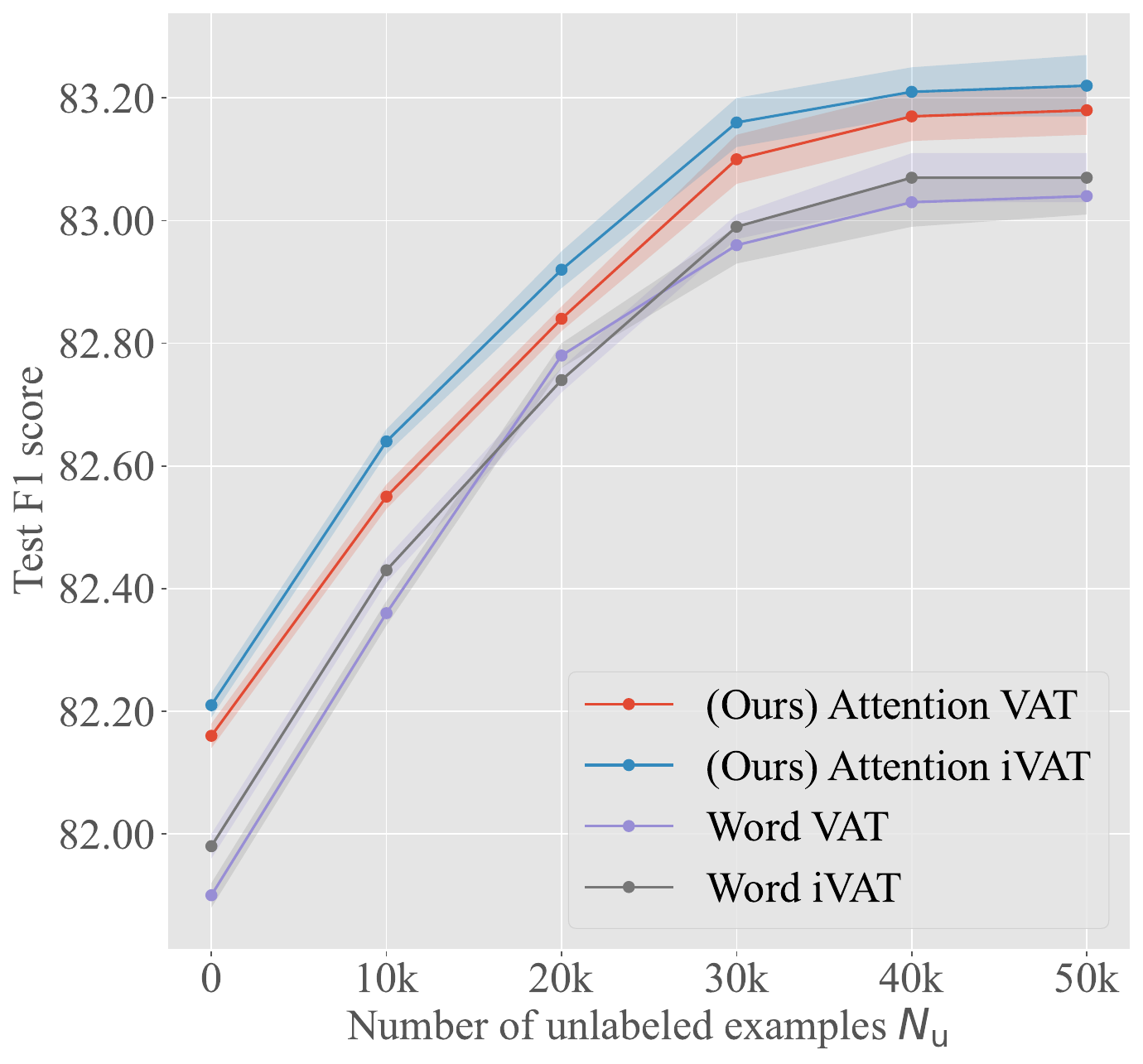}
        \subcaption{SST \\ ($N_l$ = 6,920)}
        \label{fig:unlabeled_dataset_sst}
    \end{minipage}
    \begin{minipage}{0.24\linewidth}
        \centering
        \includegraphics[width=\linewidth]{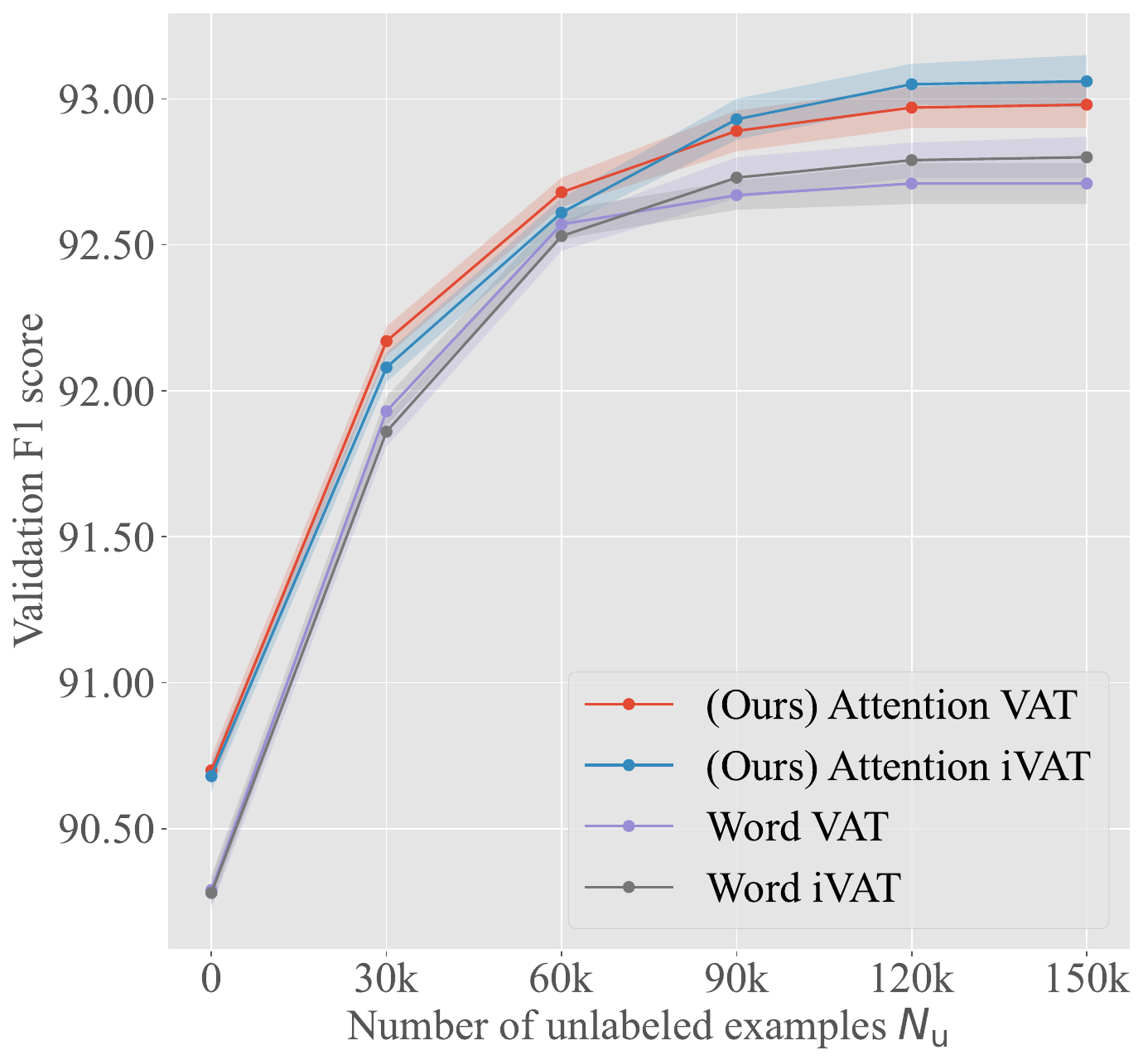}
        \includegraphics[width=\linewidth]{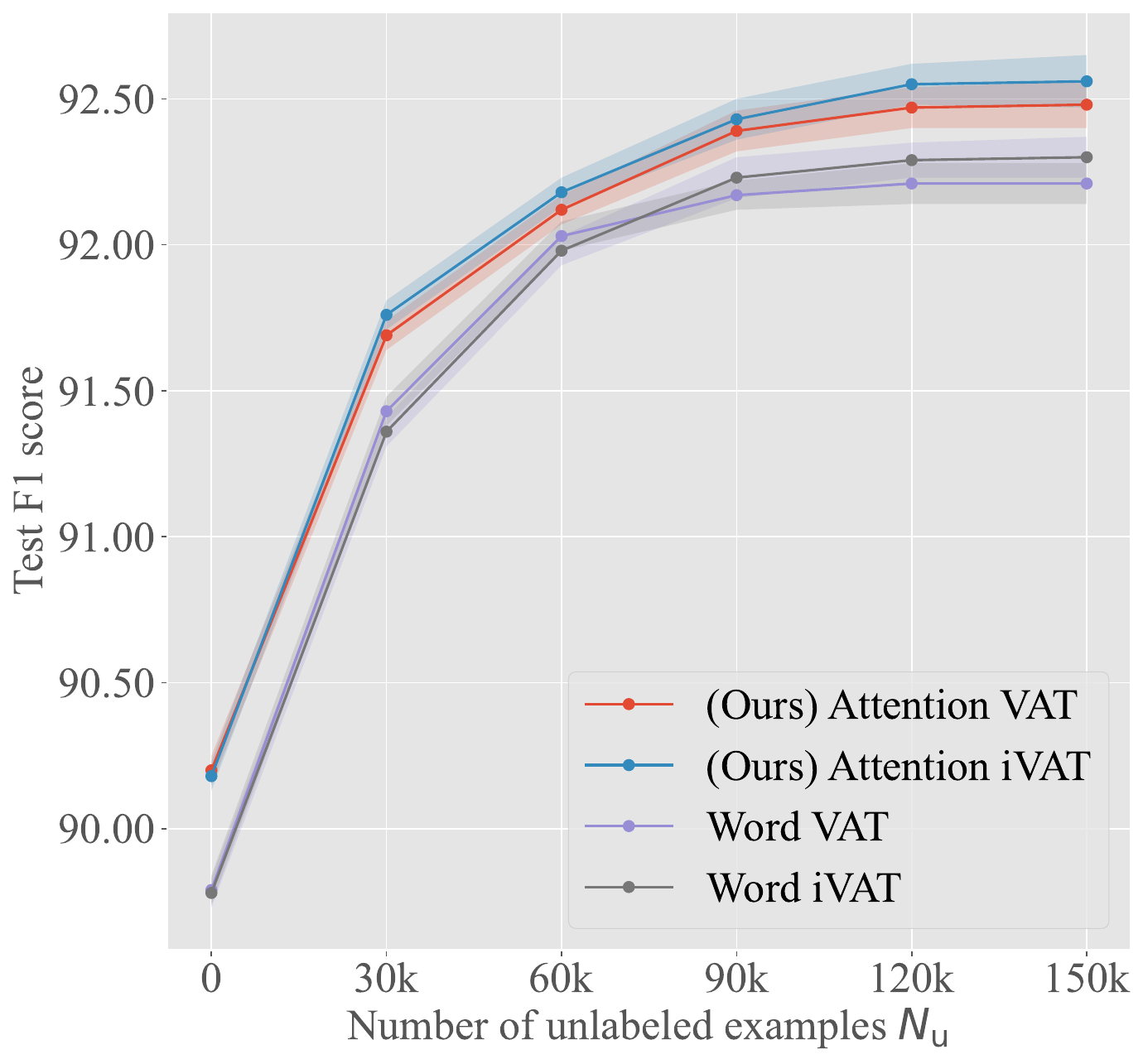}
        \subcaption{IMDB \\ ($N_l$ = 17,186)}
        \label{fig:unlabeled_dataset_imdb}
    \end{minipage}
    \begin{minipage}{0.24\linewidth}
        \centering
        \includegraphics[width=\linewidth]{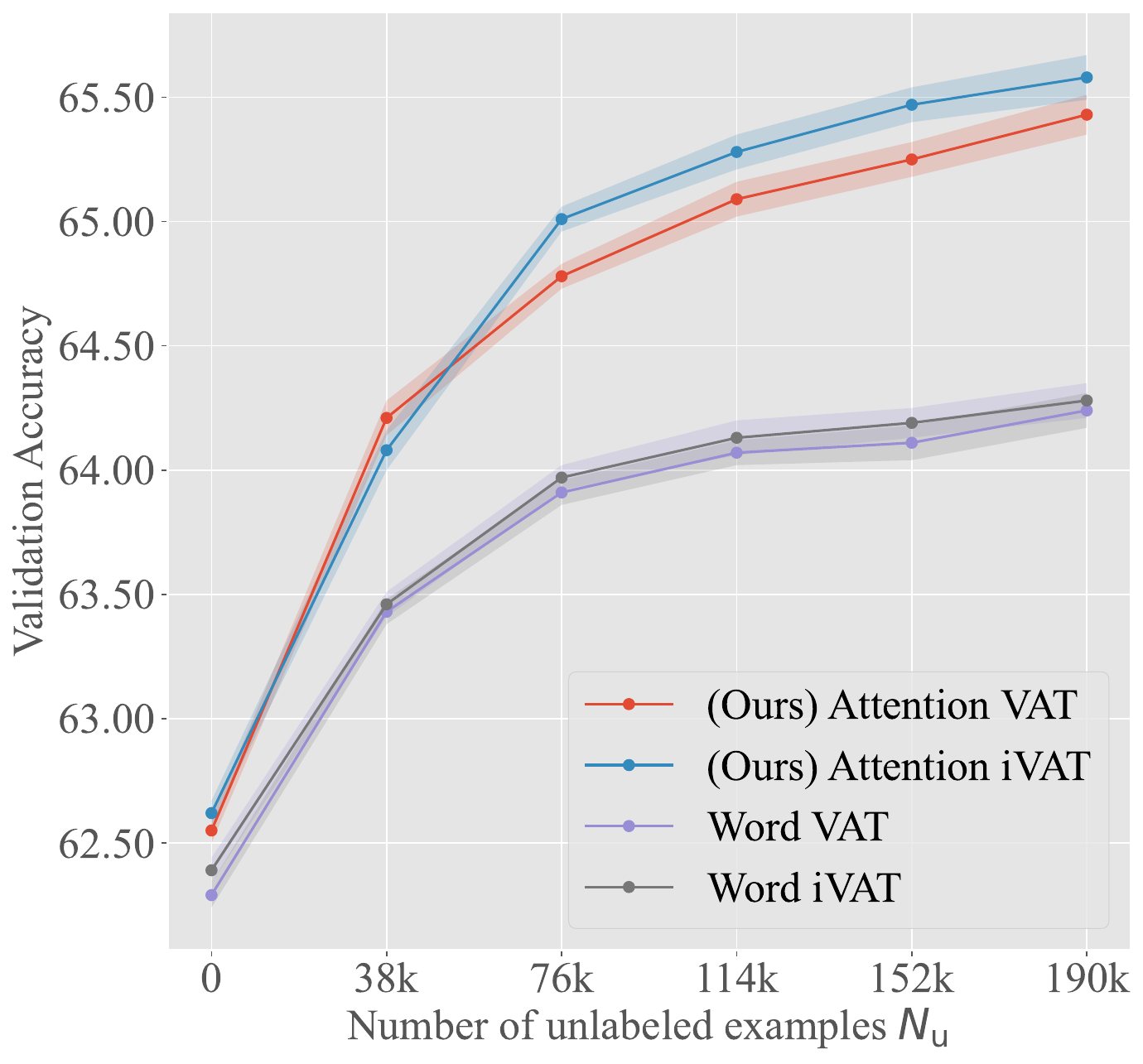}
        \includegraphics[width=\linewidth]{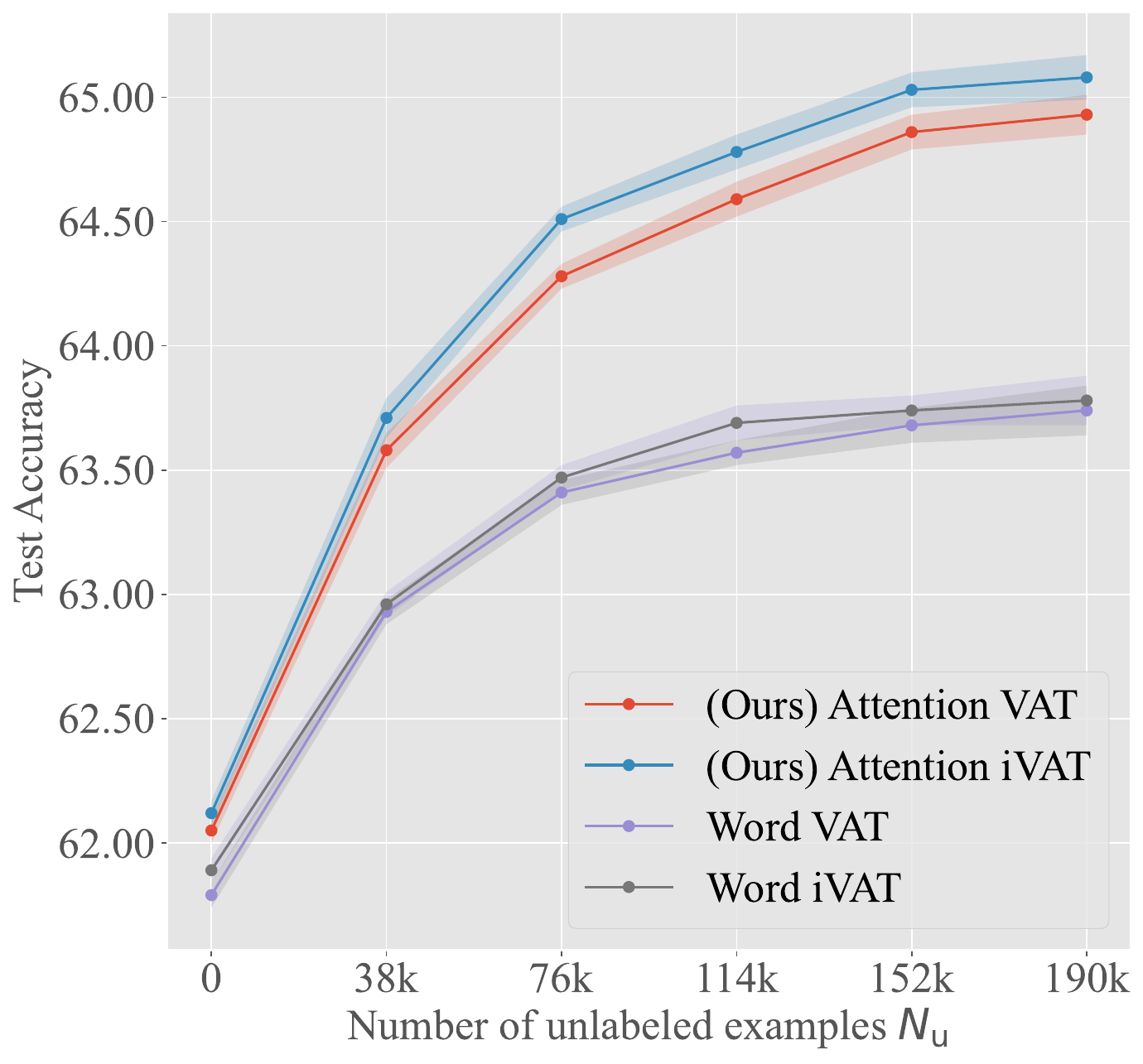}
        \subcaption{CNN news \\ ($N_l$ = 190,149)}
        \label{fig:unlabeled_dataset_cnn}
    \end{minipage}
    \begin{minipage}{0.24\linewidth}
        \centering
        \includegraphics[width=\linewidth]{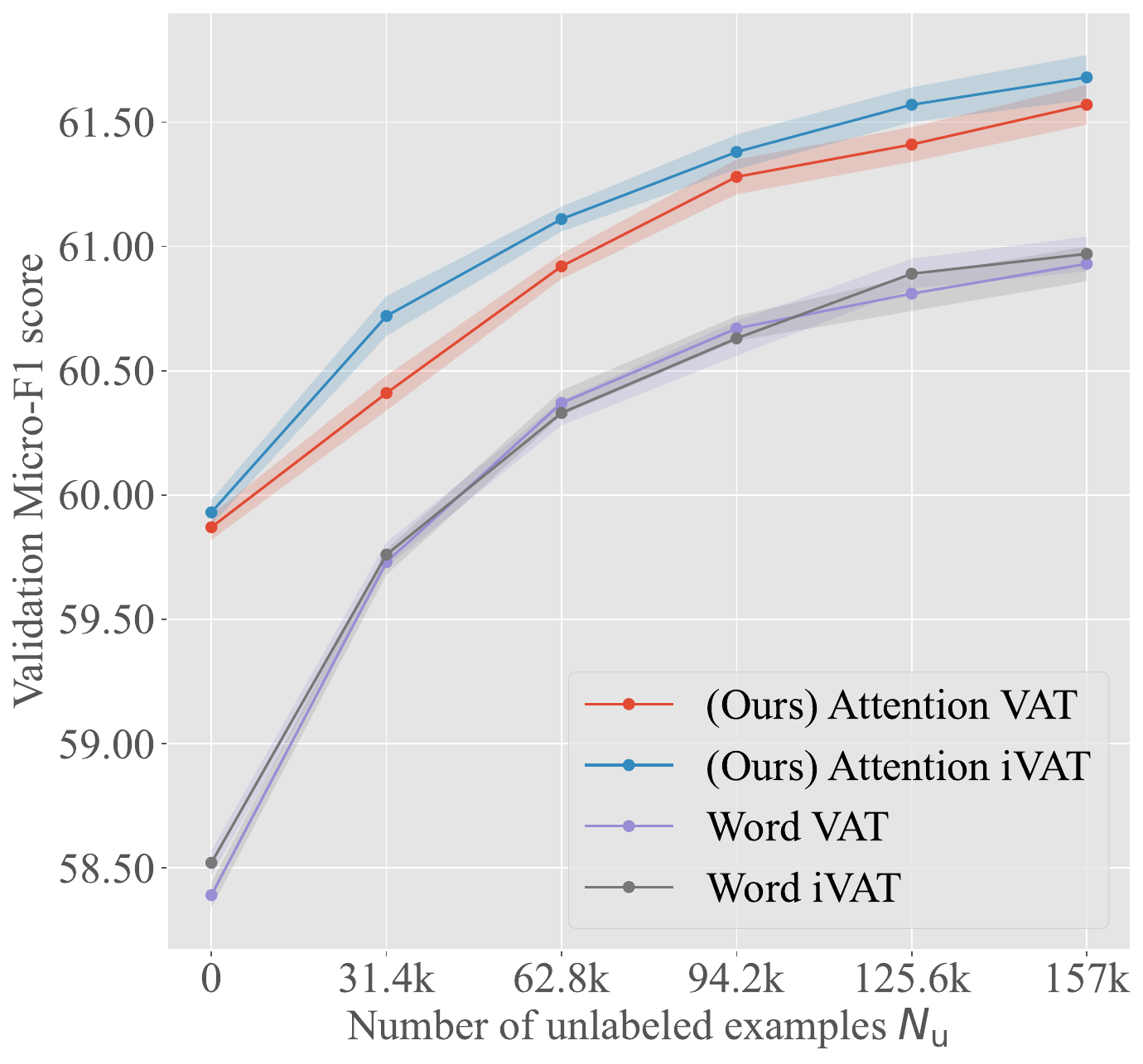}
        \includegraphics[width=\linewidth]{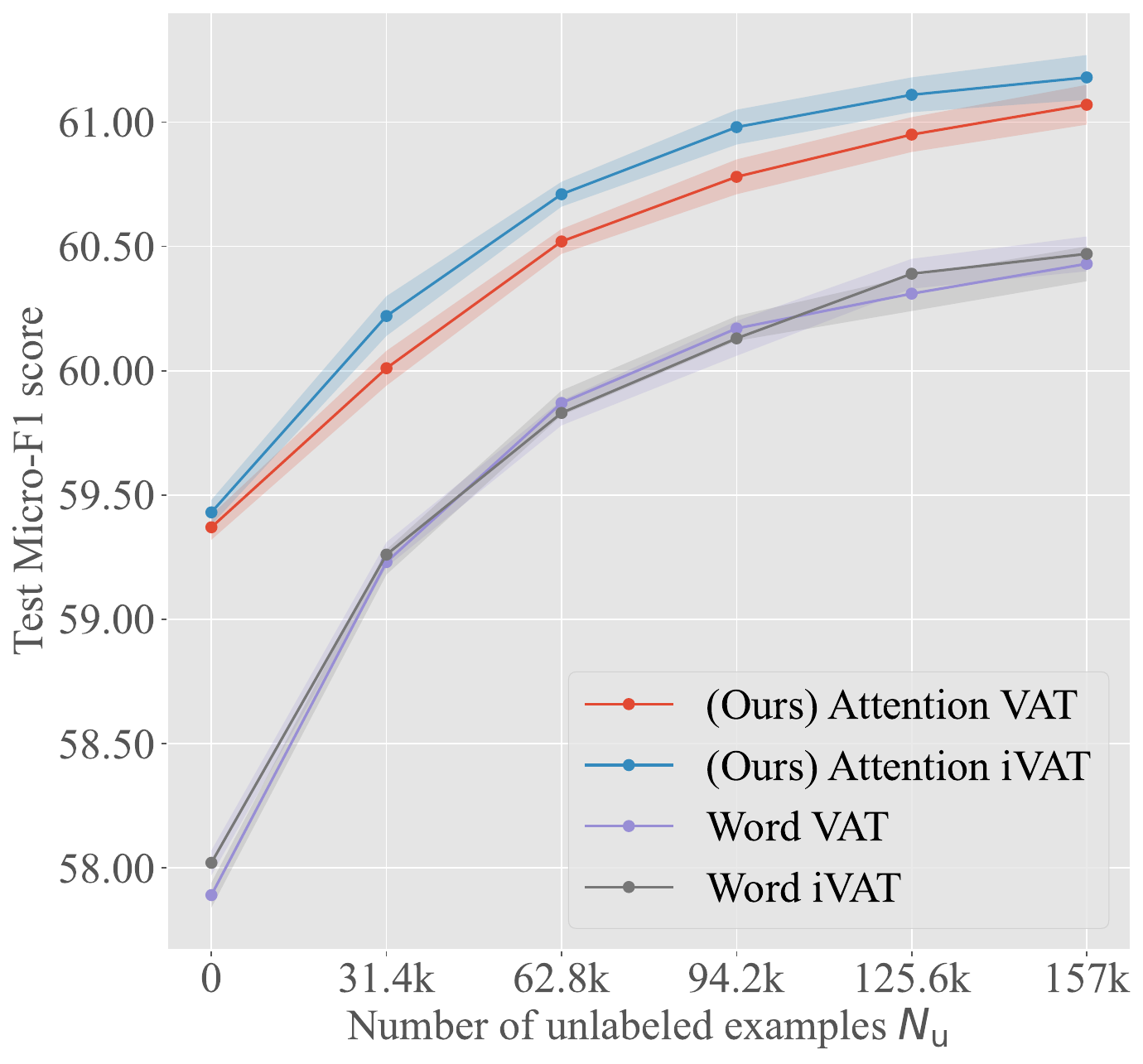}
        \subcaption{MultiNLI \\ ($N_l$ = 157,081)}
        \label{fig:unlabeled_dataset_multinli}
    \end{minipage}
     \caption{
        Model performance in validation and test score for single-sequence tasks (SST and IMDB) and pair-sequence tasks (CNN news and MultiNLI) with different amounts of unlabeled examples
    }
    \label{fig:unlabeled_dataset}
\end{figure*}

Fig.~\ref{fig:unlabeled_dataset} depicts the changes in the prediction performance with an increase in the amount of unlabeled data in the random selection strategy.
It shows that the amount of unlabeled data was a crucial factor for this performance for both validation and test data. Here, because similar results were obtained in each of the three single- and pair-sequence tasks, two for each task are shown for reason of space. The results of five experiments each with different seed values are shown with error bars.
Specifically, in the single-sequence tasks, namely SST and IMDB, the performance improved as the amount of unlabeled data increased up to approximately seven times that of the labeled data.
In the pair-sequence tasks, namely CNN news and MultiNLI, although we could only prepare up to twice the amount of unlabeled data because of the limitations of the experimental conditions as mentioned above, the accuracy also improved as the amount of data increased.
This analysis demonstrates that VAT-based techniques provide significant benefits with a large amount of unlabeled data.
Although we presented the results of four datasets (two for each dataset in the single- and pair-sequence tasks) that are representative of the six datasets used in the evaluation experiments, we found generally the same trends in the other datasets.

% \usepackage{booktabs}
\begin{table*}[t]
\centering
\resizebox{\textwidth}{!}{%
\begin{tabular}{@{}lllrrrrrr@{}}
\toprule
\multicolumn{3}{c}{Model}                                                                                                                          & \multicolumn{2}{c}{SST}                                 & \multicolumn{2}{c}{IMDB}                                & \multicolumn{2}{c}{AGNews}                              \\ \cmidrule(r){1-3} \cmidrule(lr){4-5} \cmidrule(lr){6-7} \cmidrule(l){8-9}
               & \begin{tabular}[c]{@{}l@{}}Selection \\ strategy\end{tabular} & \begin{tabular}[c]{@{}l@{}}Sentence \\ representation\end{tabular} & \multicolumn{1}{c}{F1 [\%]} & \multicolumn{1}{c}{Corr.} & \multicolumn{1}{c}{F1 [\%]} & \multicolumn{1}{c}{Corr.} & \multicolumn{1}{c}{F1 [\%]} & \multicolumn{1}{c}{Corr.} \\ \cmidrule(r){1-1} \cmidrule(lr){2-2} \cmidrule(lr){3-3} \cmidrule(lr){4-4} \cmidrule(lr){5-5} \cmidrule(lr){6-6} \cmidrule(lr){7-7} \cmidrule(lr){8-8} \cmidrule(l){9-9}
Attention VAT  & Random                                                       &                                                                    & 83.18                       & 0.898                     & 92.48                       & 0.879                     & 96.35                       & 0.903                     \\
               & ANN                                                          & BoW                                                                & 83.20                       & 0.903                     & 92.51                       & 0.884                     & 96.38                       & 0.908                     \\
               &                                                              & Avg. word embedding                                                & 83.21                       & 0.927                     & 92.50                       & 0.902                     & 96.39                       & 0.928                     \\ \cmidrule(r){1-1} \cmidrule(lr){2-2} \cmidrule(lr){3-3} \cmidrule(lr){4-4} \cmidrule(lr){5-5} \cmidrule(lr){6-6} \cmidrule(lr){7-7} \cmidrule(lr){8-8} \cmidrule(l){9-9}
Attention iVAT & Random                                                       &                                                                    & 83.22                       & 0.901                     & 92.56                       & 0.883                     & 96.48                       & 0.917                     \\
               & ANN                                                          & BoW                                                                & 83.26                       & 0.909                     & 92.59                       & 0.891                     & 96.51                       & 0.924                     \\
               &                                                              & Avg. word embedding                                                & 83.26                       & 0.941                     & 92.60                       & 0.914                     & 96.52                       & 0.957                     \\ \bottomrule
\end{tabular}%
}
\caption{
Comparison of unlabeled data selection performance in semi-supervised learning. 
}
\label{tab:semi_supervised_score_with_sampling_strategy}
\end{table*}

Table~\ref{tab:semi_supervised_score_with_sampling_strategy} compares the performance of the unlabeled data selection using the selection strategies in the semi-supervised learning process.
Both the random and ANN strategies exhibited almost the same prediction performance in our techniques. 
This result suggests that the prediction performances of our techniques were almost the same, regardless of the selection strategy used. 
However, a remarkable improvement in the correlation with the word importance was observed.
This finding is discussed in further detail in Section~\ref{sec:amount_and_selection_of_unlabeled_data}.

\section{Discussion}
    \subsection{AT and VAT for Attention Mechanisms}

Our Attention VAT/iVAT techniques logically extend our Attention AT/iAT~\cite{kitada2021attention} techniques, according to our belief that attention is more important in identifying significant words in text/document processing than the actual word embeddings. 
Our proposed Attention VAT/iVAT significantly outperformed the Word AT/iAT in supervised settings, thereby reaffirming the above assumption.
Moreover, our Attention VAT/iVAT exhibited equivalent performance to that of Attention AT/iAT, although our techniques did not use supervised labels for estimating the (virtual) adversarial perturbation. 
In such situations, AT, which can use supervised labels directly to compute the adversarial direction, is advantageous. 
However, VAT, which relies on the current estimation for the computation and does not use labels, exhibits a fully equivalent performance, with no negative evidence observed. 

In the semi-supervised settings, our VAT-based techniques further improved prediction performance and the correlation with the word importance in both the single- and pair-sequence tasks. 
Our techniques can use unlabeled data more efficiently than the conventional Word VAT, which, as well as its extension Word iVAT, have been validated extensively in semi-supervised settings.
In summary, the application of VAT to the attention mechanism, rather than word embedding, resulted in improved model robustness (i.e., improved prediction performance) and interpretability (i.e., improved agreement with manually annotated rationales).

\subsection{Amount and Selection of Unlabeled Data}\label{sec:amount_and_selection_of_unlabeled_data}

The amount and selection of unlabeled data that are added for training the model are important factors for improving the performance in semi-supervised learning.
First, we discuss the relationship between the amount of unlabeled data and the prediction performance.
Our proposed techniques improved the performance until the unlabeled data were approximately seven times larger than the labeled data in the single-sequence tasks.
Although the amount of unlabeled data was limited to the amount of training data, for the previously mentioned reasons, in the pair-sequence tasks an increase in the amount of unlabeled data sufficiently improved the prediction performance.
This success is assumed to be owing to the VAT effectively utilizing unlabeled data, which contributed to smoothing the discriminative boundaries.
It is noteworthy that the data added as unlabeled data produced a performance improvement even though they differed from the original dataset regarding type and quality.

Next, we discuss the selection strategy for unlabeled data.
It is desirable for the input spaces of the unlabeled and labeled data to be similar when computing virtual adversarial perturbations using unlabeled data.
Contrary to this empirical rule, our VAT for attention mechanisms worked effectively even when unlabeled data, from a domain different from that of the labeled data, were used.
Our VAT for attention mechanisms essentially improved the model robustness. 
This confirms that our technique yields correct inference results for various inputs and can estimate the words that are important for the prediction as clear attention weights.
As VAT computes the direction of the adversarial perturbation based on current model estimation, the addition of unlabeled data contributed to improving the performance, regardless of the selection strategy for the unlabeled data.
This is possibly why our VAT for attention mechanisms need not be careful when selecting unlabeled data.

The training with more semantically similar unlabeled data from the pool (i.e., the ANN-based selection strategy) exhibited a stronger correlation between attention to words and the gradient-based estimates of the word importance than that without this consideration (i.e., the random selection strategy).
The selection of similar unlabeled data is expected to offer certain advantages in terms of interpretability. 
As the ANN-based strategy selects similar data in a low-dimensional space, a higher correlation is anticipated.
A performance improvement can be expected if a superior selection strategy is established in the future.
Nevertheless, we emphasize that the proposed technique can improve model performance by simply adding a large unlabeled data pool to the training, without considering the similarity of the data.

\subsection{Agreement with Human Annotation}

The proposed Attention iVAT demonstrably exhibited substantially stronger agreement with human annotation than all other conventional techniques.
In this evaluation, the BERT-to-BERT model~\cite{deyoung2019eraser} performed best, but this model was the only one that was trained to predict the textual portion of the rationale/evidence in a supervised manner.
Hence, it was not possible to compare directly the models that applied our techniques.
Our training technique can indirectly achieve highly efficient rationale selection for common and practical RNN-based models without using large models such as BERT and explicit learning for the rationale selection.

Similar to our previous proposal Attention iAT, Attention iVAT incorporates a process to emphasize the difference in attention to each word better, which may have contributed to the improved performance.
Specifically, the norm of the weight of the difference in the attention of words (Eq.~(\ref{eq:normalized_attention_distance_vector})) was normalized to 1.
This process makes more effective use of the differences in attention to each word and calculates the virtual adversarial perturbations, which are determined based on more data than those in the former Attention iAT.
Therefore, we believe that Attention iVAT can generate cleaner attention, and thus, achieve a higher hard/soft rationale selection score.

\section{Conclusion}
    We have proposed novel, effective, and general-purpose semi-supervised training techniques for NLP tasks, namely Attention VAT and Attention iVAT.
These methods are extensions to a semi-supervised setting using VAT of our previous techniques, namely Attention AT and Attention iAT respectively.
Our training techniques significantly improve the prediction performance and model interpretability.
In particular, our proposed techniques are highly effective when using unlabeled data and can outperform conventional VAT-based techniques.
We confirmed that our techniques perform effectively even when using unlabeled data from a source other than the labeled data. 
That is, no careful selection (i.e., simple random selection) from the source is required.
Our techniques can easily be applied to any DNN model with attention mechanisms.
In the future, we will also investigate the effects of our techniques when using attention mechanisms based on state-of-the-art models, such as Transformer and BERT models.

\section*{Acknowledgements}
This work was partially supported by JSPS KAKENHI under Grant 21J14143.

\bibliography{references}
\bibliographystyle{acl_natbib}

\appendix
\section{Common Model Architecture}\label{sec:common_model}

\begin{figure*}[t]
    \begin{minipage}{0.37\textwidth}
        \centering
        \includegraphics[width=\linewidth]{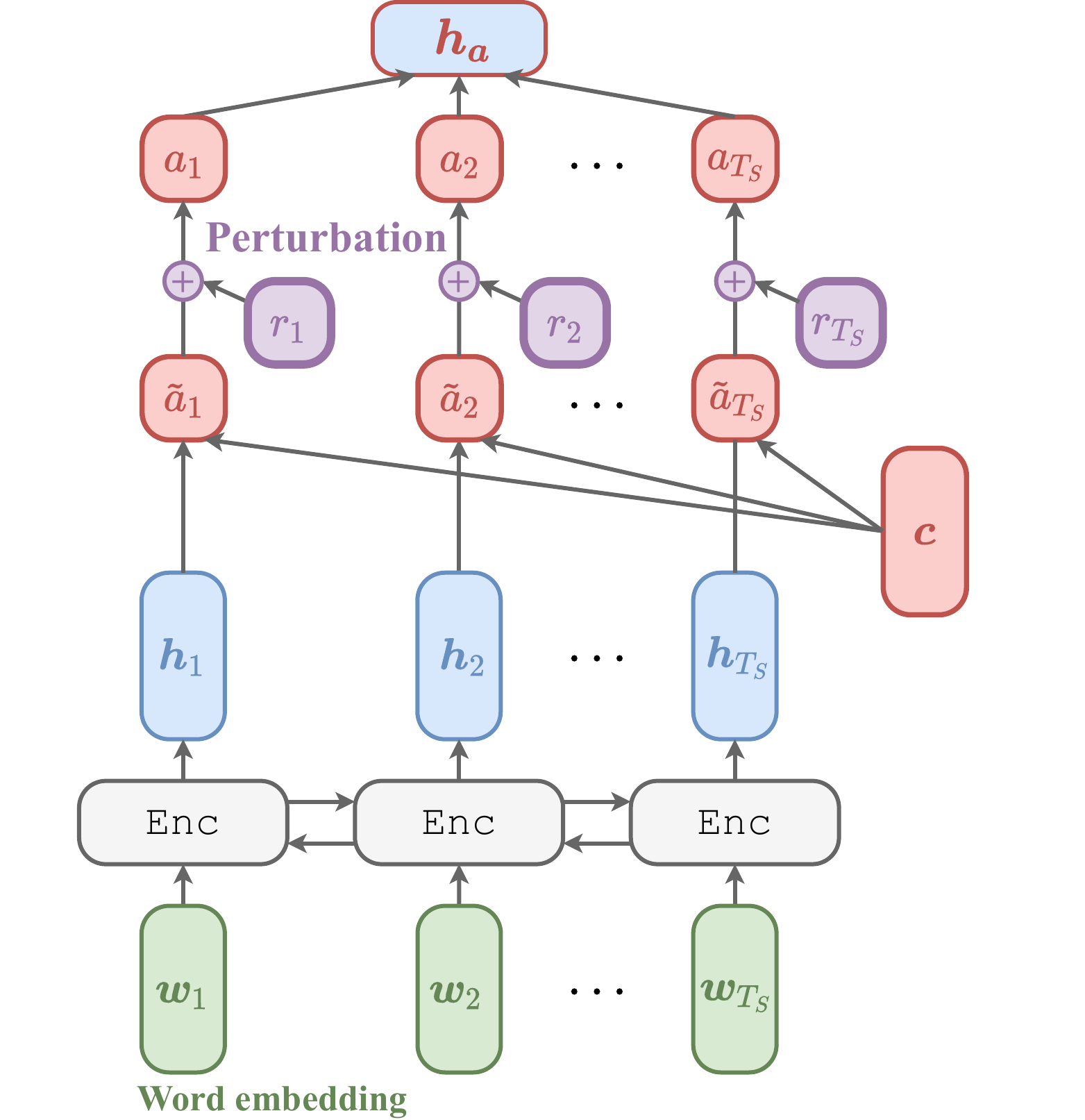}
        \subcaption{Single-sequence model}
        \label{fig:single_sequence_model}
    \end{minipage}
    \begin{minipage}{0.63\textwidth}
        \centering
        \includegraphics[width=\linewidth]{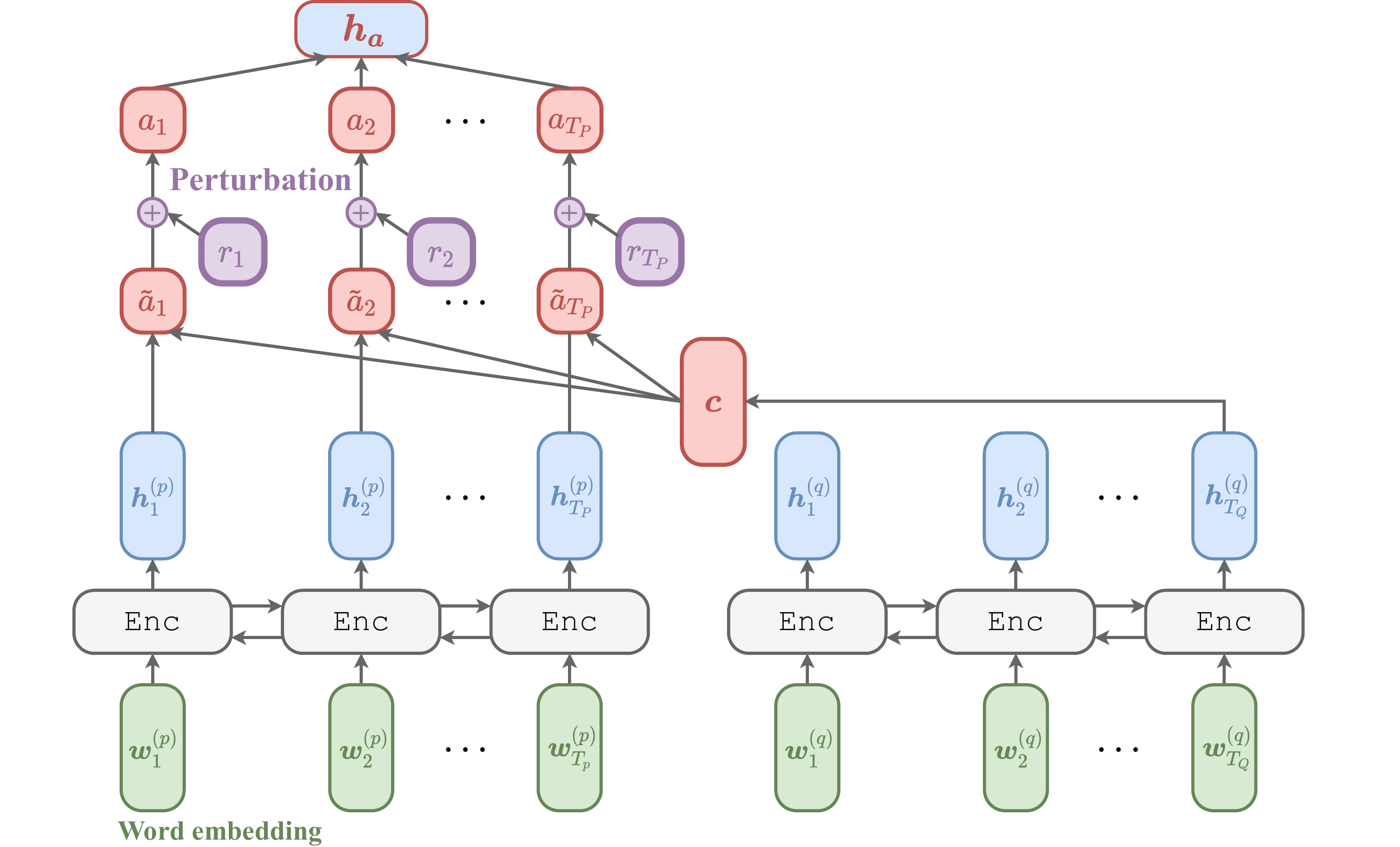}
        \subcaption{Pair-sequence model}
        \label{fig:pair_sequence_model}
    \end{minipage}
    \caption{
    Common models for applying the proposed training technique: (a) single-sequence model for the text classification task, and (b) pair-sequence model for QA and NLI tasks
    In (a), the input of the model was the word embeddings, $\{\bm{w}_1, \cdots, \bm{w}_{T_S}\}$ that were associated with the input sentence $X_S$. In (b), the inputs were the word embeddings $\{\bm{w}_{1}^{(p)}, \cdots, \bm{w}_{T_P}^{(p)}\}$ and $\{\bm{w}_{1}^{(q)}, \cdots, \bm{w}_{T_Q}^{(q)}\}$ from the two input sequences, $X_P$ and $X_Q$, respectively. These inputs were encoded into hidden states through a bidirectional encoder (\textbf{Enc}). In conventional models, the worst-case perturbation $\bm{r}$ is added to the word embeddings. In our Attention VAT and iVAT, $\bm{r}$ is computed and added to the attention score $\tilde{\bm{a}}$ to improve the prediction performance and model interpretability, even with unlabeled data.
    }
    \label{fig:model_overview}
\end{figure*}

In this appendix, we introduce the common model architecture to which our training techniques were applied. 
The common model is a practical and widely used RNN-based model, and its performance has been compared in extensive experiments focusing on attention mechanisms~\cite{jain2019attention, kitada2021attention, meister2021sparse}.
As the settings differ for single- and pair-sequence tasks, we defined a model for each task, as illustrated in Fig.~\ref{fig:model_overview}, and the details are described below.

\subsection{Model for Single-Sequence Tasks}

Fig.~\ref{fig:single_sequence_model} presents the model for the single-sequence tasks, such as text classification. 
The input of the model was a word sequence of one-hot encoding $X_S = (\bm{x}_1, \bm{x}_2, \cdots, \bm{x}_{T_S}) \in \mathbb{R}^{\abs{V} \times T_S}$, where $\abs{V}$ and $T_S$ are the vocabulary size and the number of words in the sequence, respectively.
Let $\bm{w}_t \in \mathbb{R}^{d}$ be a $d$-dimensional word embedding corresponding to $\bm{x}_t$.
Each word was represented with the word embeddings to obtain $(\bm{w}_t)_{t=1}^{T_S} \in \mathbb{R}^{d \times T_S}$.
The word embeddings were encoded with a bidirectional RNN (BiRNN)-based encoder $\mathbf{Enc}$ to obtain the $m$-dimensional hidden state: 
\begin{equation}
    \bm{h}_t = \mathbf{Enc}(\bm{w}_t, \bm{h}_{t-1}), 
\end{equation}
where $\bm{h}_0$ is the initial hidden state and it is regarded as a zero vector.
Following \citet{jain2019attention} and \citet{kitada2021attention}, we used the additive formulation of attention mechanisms~\cite{bahdanau2014neural} to compute the attention score for the $t$-th word $\tilde{a}_t$, which is defined as:
\begin{equation}
    \tilde{a}_t = \bm{c}^{\top} \mathrm{tanh}(W\bm{h}_t + \bm{b}), 
\end{equation}
where $W \in \mathbb{R}^{d' \times m}$ and $\bm{b}, \bm{c} \in \mathbb{R}^{d'}$ are the model parameters.
Subsequently, the attention weights $\bm{a} \in \mathbb{R}^T$ for all words were computed from the attention scores $\tilde{\bm{a}} = (\tilde{a}_t)_{t=1}^{T_S}$, as follows:
\begin{equation}
    \bm{a} = (a_t)_{t=1}^{T_S} = \mathrm{softmax}(\tilde{\bm{a}}).\label{eq:attention_weight}
\end{equation}
The weighted instance representation $\bm{h}_{\bm{a}}$ was calculated using the attention weights $\bm{a}$ and hidden state $\bm{h}_t$, as follows:
\begin{equation}
    \bm{h}_{\bm{a}} = \sum_{t=1}^{T_S} a_t \bm{h}_t.
\end{equation}
Finally, $\bm{h}_{\bm{a}}$ was fed to a dense layer $\mathbf{Dec}$, and the output activation function $\sigma$ was used to obtain the following predictions:
\begin{equation}
    \hat{\bm{y}} = \sigma(\mathbf{Dec}(\bm{h}_{\bm{\bm{a}}})) \in \mathbb{R}^{\abs{\bm{y}}},\label{eq:prediction}
\end{equation}
where $\sigma$ is a sigmoid function and $\abs{\bm{y}}$ is the class label set size.

\subsection{Model for Pair-Sequence Tasks}

Fig.~\ref{fig:pair_sequence_model} presents the model for pair-sequence tasks, such as QA and NLI. 
The input of the model was $X_P = (\bm{x}_t^{(p)})_{t=1}^{T_P} \in \mathbb{R}^{\abs{V} \times T_P}$ and $X_Q = (\bm{x}_t^{(q)})_{t=1}^{T_Q} \in \mathbb{R}^{\abs{V} \times T_Q}$, where $T_P$ and $T_Q$ are the number of words in each sentence.
Furthermore, $X_P$ and $X_Q$ represent the paragraph and question in QA tasks and the hypothesis and premise in NLI tasks, respectively.
We used two separate BiRNN encoders ($\mathbf{Enc}_P$ and $\mathbf{Enc}_Q$) to obtain the hidden states $\bm{h}_{t}^{(p)} \in \mathbb{R}^m$ and $\bm{h}_{t}^{(q)} \in \mathbb{R}^m$:
\begin{equation}
    \bm{h}_{t}^{(p)} = \mathbf{Enc}_P(\bm{w}_{t}^{(p)}, \bm{h}_{t-1}^{(p)});
\end{equation}
\begin{equation}
    \bm{h}_{t}^{(q)} = \mathbf{Enc}_Q(\bm{w}_{t}^{(q)}, \bm{h}_{t-1}^{(q)}),
\end{equation}
where $\bm{h}_{0}^{(p)}$ and $\bm{h}_{0}^{(q)}$ are the initial hidden states, and they are regarded as zero vectors.
Subsequently, we computed the attention weight $\tilde{a}_t$ of each word of $X_P$ as follows:
\begin{equation}
    \tilde{a}_t = \bm{c}^{\top} \mathrm{tanh}(\bm{W}_1\bm{h}_t^{(p)} + \bm{W}_2\bm{h}_{T_Q}^{(q)} + \bm{b}),
\end{equation}
where $\bm{W}_1 \in \mathbb{R}^{d' \times m}$ and $\bm{W}_2 \in \mathbb{R}^{d' \times m}$ denote the projection matrices, and $\bm{b}, \bm{c} \in \mathbb{R}^{d'}$ are the parameter vectors.
Similar to Eq.~(\ref{eq:attention_weight}), the attention weight $a_t$ could be calculated from $\tilde{a}_t$.
The presentation was obtained from the sum of the words in $X_P$.
\begin{equation}
    \bm{h}_{\bm{a}} = \sum_{t=1}^{T_P} a_t \bm{h}_t^{(p)}
\end{equation}
was fed to a \textbf{Dec}, following which a softmax function was used as $\sigma$ to obtain the prediction (in the same manner as in Eq.~(\ref{eq:prediction})).

\subsection{Model Training with Attention Mechanisms}

Let $X_{\tilde{\bm{a}}}$ be an input sequence with an attention score $\tilde{\bm{a}}$, where $\tilde{\bm{a}}$ is a concatenated attention score for all $t$. 
The conditional probability of the class $\bm{y}$ was modeled as $p(\bm{y} \vert X_{\tilde{\bm{a}}}; \bm{\theta})$, where $\bm{\theta}$ represents all model parameters.
We minimized the following negative log-likelihood as a loss function for the model parameters to train the model:
\begin{equation}
    \mathcal{L}(X_{\tilde{\bm{a}}}, \bm{y}; \bm{\theta}) = - \log{p(\bm{y} \vert X_{\tilde{\bm{a}}}; \bm{\theta})}.
\end{equation}

\section{Implementation Details}\label{sec:implementation_details}

We implemented all of the training techniques using the AllenNLP library with Interpret~\cite{gardner2018allennlp, Wallace2019AllenNLP}.
We evaluated the test set only once in all experiments.
The experiments were conducted on an Ubuntu PC with a GeForce GTX 1080 Ti GPU.
Our implementation is based on the one published by \citet{kitada2021attention}.\footnote{\url{https://github.com/shunk031/attention-meets-perturbation}}
Note that the model that was used in our experiment had a small number of parameters compared to recent models; therefore, the execution speed of the model was also fast.

\subsection{Supervised Classification Task}

We used pretrained fastText~\cite{bojanowski2017enriching} word embedding with 300 dimensions and a one-layer bi-directional long short-term memory (LSTM)~\cite{hochreiter1997long} as the encoder with a hidden size of 256 for the supervised settings.
We used a sigmoid function as the output activation function.
All models were regularized using $L_2$ regularization ($10^{-5}$) that was applied to all of the parameters.
We trained the model using the maximum likelihood loss and the Adam optimizer~\cite{kingma2014adam} with a learning rate of 0.001.
All of the experiments were conducted at $\lambda = 1$.

We searched for the best hyperparameter $\epsilon$ from $[0.01, 30.00]$ following~\citet{kitada2021attention}.
The Allentune library~\cite{dodge2019show} was used to adjust $\epsilon$, and we decided on the value of the hyperparameter $\epsilon$ based on the validation score.

\subsection{Semi-supervised Classification Task}

We used the same pretrained fastText and encoder for the semi-supervised settings.
We again used the Adam optimizer~\cite{kingma2014adam} with the same learning rate as that in the supervised classification task.
The same hyperparameter search was performed as in the supervised settings.
All of the experiments were conducted at $\lambda = 1$ in the semi-supervised settings.

We also performed the same preprocessing according to \citet{chen2020seqvat} using the same unlabeled data.
We determined the amount of unlabeled data $N_{\mathrm{ul}}$ based on the validation score for each benchmark dataset.
We reported the test score of the model with the highest validation score.
\section{Details of Tasks and Dataset}\label{sec:appendix_tasks_and_dataset}

\subsection{Single-Sequence Task}

SST~\cite{socher2013recursive}\footnote{\url{https://nlp.stanford.edu/sentiment/trainDevTestTrees_PTB.zip}} was used to ascertain the positive or negative sentiment the a sentence.
IMDB Large Movie Reviews (IMDB)~\cite{maas2011learning}\footnote{\url{https://s3.amazonaws.com/text-datasets/imdb_full.pkl}}$^{,}$\footnote{\url{https://s3.amazonaws.com/text-datasets/imdb_word_index.json}} was used to identify positive or negative sentiments from movie reviews.
AG News (AGNews)~\cite{zhang2015character}\footnote{The dataset can be found on Xiang Zhang's \href{https://drive.google.com/drive/u/0/folders/0Bz8a_Dbh9Qhbfll6bVpmNUtUcFdjYmF2SEpmZUZUcVNiMUw1TWN6RDV3a0JHT3kxLVhVR2M}{Google Drive}.} was used to identify the topic of news articles as either the world  (set as a negative label) or business (set as a positive label).

\subsection{Pair-Sequence Task}

CNN news article corpus (CNN news)~\cite{hermann2015teaching}\footnote{The dataset can be found on Deep Mind Q\&A \href{https://drive.google.com/uc?export=download&id=0BwmD_VLjROrfTTljRDVZMFJnVWM}{Google Drive}.} was used to identify the answer entities from a paragraph.
SNLI~\cite{bowman2015large}\footnote{\url{https://nlp.stanford.edu/projects/snli/snli_1.0.zip}} was used to determine whether a hypothesis sentence entailed, contradicted, or was neutral regarding a given premise sentence.
Multi-Genre NLI (MultiNLI)~\cite{williams2017broad}\footnote{\url{https://www.nyu.edu/projects/bowman/multinli/multinli_1.0.zip}} used the same format as SNLI and was a comparable in size, but it included a more diverse range of text, as well as an auxiliary test set for cross-genre transfer evaluation.
\section{Details of Evaluation Criteria}

\subsection{Correlation between Attention and Gradient-based Word Importance}\label{sec:appendix_correlation_attention_gradient}

We computed how the attention weighted obtained through our VAT-based technique agree with the importance of words calculated by gradients~\cite{simonyan2013deep}.
This evaluation follows \citet{jain2019attention}.
The correlation $\tau_g$ is defined from the attention $\bm{a} \in \mathbb{R}^T$ and the gradient-based word importance $\bm{g} \in \mathbb{R}^T$ as follows:
\begin{equation}
    \tau_g = \mathrm{PearsonCorr}(\bm{a}, \bm{g}).
\end{equation}
The gradient-based word importance $\bm{g} = (g_t)_{t=1}^{T}$ is calculated as follows:
\begin{equation}
    g_t = \lvert\sum_{i=1}^{\abs{V}} \mathbbm{1} [X_{it} = 1] \frac{\partial \bm{y}}{\partial X_{it}} \rvert, \forall t \in [1, T],
\end{equation}
where $X_{it}$ is the $t$-th one-hot encoded word for the $i$-th vocabulary in $X \in R^{\abs{V} \times T}$, and $T$ is the number of words in the sequence.

\end{document}